\documentclass[review]{elsarticle}

\usepackage{amsmath}
\usepackage{amsfonts}
\usepackage{amssymb}
\usepackage{amsthm}
\usepackage{bm}
\usepackage{indentfirst}
\usepackage{graphicx}
\usepackage{subfigure}
\usepackage{array}
\usepackage{fullpage}

\DeclareMathAlphabet{\mathsfsl}{OT1}{cmss}{m}{sl}

\newcommand{\PreserveBackslash}[1]{\let\temp=\\#1\let\\=\temp}
\newcolumntype{C}[1]{>{\PreserveBackslash\centering}p{#1}}
\newcolumntype{R}[1]{>{\PreserveBackslash\raggedleft}p{#1}}
\newcolumntype{L}[1]{>{\PreserveBackslash\raggedright}p{#1}}

\numberwithin{equation}{section}
\newtheorem{thm}{Theorem}[section]

\theoremstyle{definition}

\usepackage{anyfontsize}
\usepackage{enumerate}
\usepackage{esint}
\usepackage{etoolbox}
\usepackage{isomath}
\usepackage{mathrsfs}
\usepackage{mathtools}
\usepackage{multicol}
\usepackage{pgfplots}
\usepackage{scalerel}
\usepackage{stackengine}
\usepackage{tabulary}
\usepackage{tikz}

\makeatletter
\newcommand*\bdot{\mathpalette\bdot@{.65}}
\newcommand*\bdot@[2]{\mathbin{\vcenter{\hbox{\scalebox{#2}{$\m@th#1\bullet$}}}}}
\makeatother

\makeatletter
\newcommand*\bddot{\mathpalette\bddot@{.65}}
\newcommand*\bddot@[2]{\mathbin{\vcenter{\hbox{\scalebox{#2}
    {$\m@th#1\smash{{}_{\bullet}^{\bullet}}$}}}}}
\makeatother

\newcommand{\circled}[2][]{%
  \tikz[baseline=(char.base)]{%
    \node[shape = circle, draw, inner sep = .5pt]
    (char) {\phantom{\ifblank{#1}{#2}{#1}}};%
    \node at (char.center) {\makebox[0pt][c]{#2}};}}
\robustify{\circled}

\makeatletter
\newcommand{\opnorm}{\@ifstar\@opnorms\@opnorm}
\newcommand{\@opnorms}[1]{%
  \left|\mkern-1.5mu\left|\mkern-1.5mu\left|
   #1
  \right|\mkern-1.5mu\right|\mkern-1.5mu\right|
}
\newcommand{\@opnorm}[2][]{%
  \mathopen{#1|\mkern-1.5mu#1|\mkern-1.5mu#1|}
  #2
  \mathclose{#1|\mkern-1.5mu#1|\mkern-1.5mu#1|}
}
\makeatother

\stackMath
\newcommand\reallywidecheck[1]{%
\savestack{\tmpbox}{\stretchto{%
  \scaleto{%
    \scalerel*[\widthof{\ensuremath{#1}}]{\kern-.6pt\bigwedge\kern-.6pt}%
    {\rule[-\textheight/2]{1ex}{\textheight}}
  }{\textheight}%
}{0.5ex}}%
\stackon[1pt]{#1}{\scalebox{-1}{\tmpbox}}%
}
\usepackage[utf8]{inputenc}
\usepackage{bm}
\usepackage{mathrsfs}
\usepackage{amsmath}
\usepackage{amsfonts}
\usepackage{amsthm}
\usepackage{algorithm}
\usepackage[noend]{algpseudocode}
\usepackage{color}
\usepackage{setspace}
\usepackage{float}
\usepackage{import}
\usepackage{url}
\usepackage{multicol}
\usepackage{subfigure}
\biboptions{sort&compress}
\usepackage[top=1in,bottom=1in,left=1in,right=1in]{geometry}
\usepackage{amsmath,amssymb,amsthm,bbm,mathtools,bm}
\usepackage{multicol}
\usepackage{multirow}
\usepackage{graphicx}
\usepackage{paralist}

\definecolor{officegreen}{rgb}{0.0, 0.5, 0.0}

\newcommand{\mcU}{\mathcal{U}}

\newcommand{\mcB}{\mathcal{B}}

\newcommand{\mcF}{\mathcal{F}}

\newcommand{\mcN}{\mathcal{N}}
\newcommand{\mcL}{\mathcal{L}}

\newcommand{\mcA}{\mathcal{A}}

\newcommand{\mbR}{\mathbb{R}}
\newcommand{\mbRd}{{\mathbb{R}^d}}

\def \bb{\mathbf{b}}

\def \fb{\mathbf{f}}
\def \ub{\mathbf{u}}

\def \wb{\mathbf{w}}
\def \vb{\mathbf{v}}
\def \xb{\mathbf{x}}

\def \yb{\mathbf{y}}

\def \kb{\mathbf{k}}
\def \hb{\mathbf{h}}
\def \cb{\mathbf{c}}
\def \pb{\mathbf{p}}
\def \qb{\mathbf{q}}

\def \real{\mathbb{R}}

\begin{document}

\begin{frontmatter}

\title{Nonlocal Kernel Network (NKN): a Stable and Resolution-Independent Deep Neural Network}

\address[yy]{Department of Mathematics, Lehigh University, Bethlehem, PA}
\address[md]{Computational Science and Analysis, Sandia National Laboratories, Livermore, CA}
\address[tg]{IBM Research, Yorktown Heights, NY}
\address[ss]{Center for Computing Research, Sandia National Laboratories, Albuquerque, NM}

\author[yy]{Huaiqian You}\ead{huy316@lehigh.edu}
\author[yy]{Yue Yu\corref{cor1}}\ead{yuy214@lehigh.edu}
\author[md]{Marta D'Elia}\ead{mdelia@sandia.gov}
\author[tg]{Tian Gao}\ead{Tgao@us.ibm.com}
\author[ss]{Stewart Silling}\ead{sasilli@sandia.gov}
\cortext[cor1]{Corresponding author}


\begin{abstract}
Neural operators \cite{lu2019deeponet,lu2021learning,li2020neural,li2020multipole,li2020fourier} 
have recently become popular tools for designing solution maps between function spaces in the form of neural networks. Differently from classical scientific machine learning approaches that learn parameters of a known partial differential equation (PDE) for a single instance of the input parameters at a fixed resolution, neural operators approximate the solution map of a family of PDEs \cite{kovachki2021neural,lu2021comprehensive}. 
Despite their success, the uses of neural operators are so far restricted to relatively shallow neural networks and confined to learning hidden governing laws. In this work, we propose a novel nonlocal neural operator, which we refer to as nonlocal kernel network (NKN), that is resolution independent, characterized by deep neural networks, and capable of handling a variety of tasks such as learning governing equations and classifying images.
Our NKN stems from the interpretation of the neural network as a discrete nonlocal diffusion reaction equation that, in the limit of infinite layers, is equivalent to a parabolic nonlocal equation, whose stability is analyzed via nonlocal vector calculus. The resemblance with integral forms of neural operators allows NKNs to capture long-range dependencies in the feature space, while the continuous treatment of node-to-node interactions makes NKNs resolution independent. The resemblance with neural ODEs, reinterpreted in a nonlocal sense, and the stable network dynamics between layers allow for generalization of NKN's optimal parameters from shallow to deep networks. This fact enables the use of shallow-to-deep initialization techniques \cite{ruthotto2019deep}.  
Our tests show that NKNs outperform baseline methods in both learning governing equations 
and image classification tasks and generalize well to different resolutions and depths.
\end{abstract}


\end{frontmatter}


\tableofcontents

\section{Introduction}

During the last 20 years there has been a lot of progress in the design of neural networks (NNs); however, their employment in scientific machine learning with the purpose of learning hidden physics of complex system is relatively recent. In this work, we consider the problem of designing optimal deep NNs for learning tasks such as identifying unknown governing laws and classifying images. To achieve this goal, we pursue a new network architecture that 1) is guaranteed to be independent of the input resolution, 2) is stable in the limit of deep layers, and 3) considers long-range interactions in the feature space (i.e. node-to-node interactions).
Among relevant works that use NNs with the purpose of learning governing laws, we mention physics-informed NNs \cite{raissi2019physics} where the solution of a {\it (partially) known} partial differential equation (PDE) is modeled by a deep NN whose weights and biases are learned together with the PDE's unknown parameters. More recently, the use of NNs has been extended to learning maps between inputs of a dynamical system and its state, so that the network is a surrogate for a solution operator and it can be referred to as {\it neural operator} \cite{lu2019deeponet,lu2021learning,li2020neural,li2020multipole,li2020fourier}. This approach finds applicability when constitutive laws are unknown or when the presence of high degrees of heterogeneity makes classical, PDE models inaccurate. Relevant works in this direction are the graph kernel network (GKN) architecture \cite{li2020neural,li2020multipole} (also known as the first form of a integral neural operator), the Fourier neural operator (FNO) architecture \cite{li2020fourier}, and the DeepONet architecture \cite{lu2019deeponet,lu2021learning}. 


We briefly discuss the implications of the properties 1)--3). Being resolution independent implies that the accuracy of the prediction is invariant with respect to the resolution of input parameters such as loadings and material properties. This fact is in stark contrast with classical finite-dimensional approaches which build NN models between finite-dimensional Euclidean spaces, so that their accuracy is tied to the input's resolution \cite{guo2016convolutional,zhu2018bayesian,adler2017solving,bhatnagar2019prediction,khoo2021solving}. Furthermore, being generalizable with respect to different input parameter instances means that once the neural operator is trained, solving for a new instance of the input parameter only requires a forward pass of the network. This property is in contrast with traditional PDE-constrained optimization techniques \cite{de2015numerical} and some NN models which directly parameterize the solution \cite{raissi2019physics,weinan2018deep,bar2019unsupervised,smith2020eikonet,pan2020physics}, as these methods only approximate the solution for a {\it single instance of the input}. 

Being stable in the limit of deep layers is particularly important when the complexity of the problem at hand requires deep networks to achieve a desired prediction accuracy. This is the case in tasks such as learning governing equations of complex systems (as will be clear later on in the paper) and in image classification tasks. The lack of stability occurs in different forms with error stagnation and vanishing gradients being the most common. Being able to guarantee that, by construction, the network architecture will not incur any of these issues, warrants robustness and trustability of the surrogate.

Enabling long-range interactions within the set of nodes, or, in other words, node-to-node interactions, makes a neural operator particularly suitable for identifying physical laws for highly-heterogeneous physical systems thanks to the fact that the architecture can explore interactions in the feature space and, as testified by several examples in the literature (see, e.g., convolutional NNs \cite{avelar2019discrete} where parts of the node set interact via convolutional operators), make the architecture suitable for image processing tasks.

We point out that achieving these properties is not new and there are several examples in the literature of architectures that achieve some of the properties above. What is lacking, and what we achieve in this paper, is the design of a network architecture that embeds all properties 1)--3). Below, we provide a concise summary of architectures that feature some of our desired properties and highlight their advantages and limitations. In convolutional NNs (CNNs) \cite{avelar2019discrete,lefkimmiatis2017non,o2015introduction}, the interaction of nodes within network layers is achieved via convolutional operators and makes the network particularly suitable for image processing tasks, thanks to its ability to learn complex and nonlinear dependencies in the feature space. In a similar manner, graph neural networks (GNNs) take into account long-range interactions via graph operators \cite{gu2020implicit,iakovlev2020learning,poli2019graph,xhonneux2020continuous}. Despite their success, the applicability of these networks can be hindered by the following issues. First, in both CNNs and GNNs the connection between nodes is achieved via discrete operators, making the resulting network resolution dependent which limits its generalizability and practicability. Second, during the training of GNNs, slow convergence or even divergence may occur, especially in the limit of deep layers \cite{tao2018nonlocal}. 

To circumvent the first issue above while maintaining node-to-node interactions so to achieve resolution-independent networks, a few works in the literature propose to connect nodes within layers by continuous operators \cite{alet2019graph,haber2018learning,li2020neural} and treat the set of nodes as a continuum so that the value of the network at each layer is a continuous function of a ``space'' variable (the nodes) and may be interpreted as the state of a system over the space domain (i.e. the continuum feature space). Among these works, the graph kernel network (GKN) approach, proposed in \cite{li2020neural} can be interpreted as a continuous version of a GNN or of the nonlocal NN introduced in \cite{Wang2018nonlocal}. However, while achieving properties 1) and 3), this architecture may feature instabilities in the limit of deep layers, hence failing to achieve property 2). Despite this, GKNs have been successfully used in PDE learning tasks in the context of Darcy's flow and Navier-Stokes equations \cite{li2020neural,li2020multipole}. 

With the purpose of improving the stability in GNNs,  Tao et al \cite{tao2018nonlocal} proposed a nonlocal NN (NNN) whose network update is characterized by a nonlocal discrete operator \cite{Du2012} that allows one to reinterpret the network as a discretization of a nonlocal diffusion equation, for which stability results are available. This network architecture achieves properties 2) and 3); however, by treating the interactions within nodes in a discrete manner, this architecture is not resolution independent, hence failing at achieving property 1). Moreover, as opposed to GKN's where the integral operators are parameterized, in this architecture the integral operators are defined in advance, so that the only parameters to be learned are the weights of the network. This reduces the descriptive power of these operators that may fail in complex learning tasks, such as in PDE learning problems. In fact, in \cite{tao2018nonlocal}, NNNs were employed only in image classification tasks, where they outperformed standard ResNet approaches by adding NNN's network updates within ResNet layers.

The architecture we propose can be interpreted as a combination of GKNs and the continuous counterpart of NNN, so that we inherit the advantages of both architectures and circumvent their limitations. Specifically, we treat node-to-node interactions continuously by means of an integral operator that is equivalent to a nonlocal diffusion-reaction operator. As such, our network is guaranteed to be resolution independent and stable even in the deep layer limit. The latter claim is supported by the nonlocal vector calculus theory that allows us to establish stability properties via variational arguments. Our proposed architecture, which we refer to as nonlocal kernel networks (NKN), outperforms GKNs, FNO and NNNs in both PDE learning and image classification tasks. 
The interpretation of NKNs as a parabolic nonlocal equation also allows us to consider the deep network limit and to exploit initialization methods recently developed for deep CNNs \cite{haber2018learning}. Specifically, we consider a shallow-to-deep initialization technique \cite{haber2018learning,modersitzki2009fair} where optimal parameters learned on shallow networks are considered as (quasi-optimal) initial guesses for deeper networks. The use of NKNs updates within CNNs augmented with the shallow-to-deep technique outperforms standard CNN approaches in image classification tasks.

We summarize our major contributions below.
\begin{enumerate}
    \item We introduce a novel  deep neural network based on nonlocal theory, referred to as NKN, that models the feature space continuously, by means of integral operators acting on the node domain.
    \item By identifying layers with time instants, NKNs can be interpreted as discretized nonlocal time-dependent diffusion-reaction equations and their limit as the number of layers goes to infinity is a nonlocal parabolic equation. Consequently, by means of the nonlocal vector calculus we can guarantee the stability of NKNs. 
    \item The interpretation of NKNs as a diffusion-reaction equation also allows for accelerated learning techniques for deep networks, such as the shallow-to-deep technique \cite{haber2018learning}, for which optimal parameters of shallow networks are used as initial guesses of deeper networks. 
    \item When applied to the task of learning governing equations, NKNs' accuracy is independent of the resolution of the input so that different input discretizations can be handled in an equally accurate manner.
    \item When applied to image classification tasks, NKNs not only are stable in the deep network limit but also enable classification of high-resolution images trained with low-resolution images and vice versa.
    \item NKNs are general and flexible with respect to tasks: not only do they handle both learning governing equations and image classification tasks, but, in both cases they outperform baseline methods.
\end{enumerate}

\paragraph{Paper Outline} In Section \ref{sec:background} we introduce three network architectures that inspired our work and highlight their advantages and limitations. In Section \ref{sec:nkn} we introduce NKNs and recall fundamental concepts of the nonlocal vector calculus. With these analysis tools, we then prove the stability of NKNs and describe efficient initialization techniques. In Section \ref{sec:experiments} we report several experiments that illustrate the efficacy of our network in comparison with baseline networks such as GKNs, FNOs, NNNs, and multiscale CNNs. Specifically, we consider two examples in the context of learning hidden governing laws (using as a reference the Poisson and Darcy equations) and two image data sets for which we perform image classification. In Section \ref{sec:conclusion} we provide a summary of our achievements and concluding remarks. In \ref{sec:newapp_pde}, additional numerical results are provided.

\section{Background and Related Work} \label{sec:background}

This section provides the necessary background for the rest of the paper and it is organized in two parts. First, we review three approaches recently proposed in the literature that inspired the proposed NKN and highlight their benefits and limitations, as summarized in Table \ref{tab:comparison}. NKNs are designed in such a way that all the benefits of these approaches are preserved, while limitations are overcome.

\begin{table}
\begin{center}
{\small\begin{tabular}{ c | c | c | c | c | c | c }
\hline
 Model & PDE &  Image  &Continuous in& Resolution  & Stability in &Ref\\
       & Learning    & Classification &  Depth (Time)          & Independence & Deep Networks&\\ \hline
GKN and FNO & \checkmark & -- & -- & \checkmark & -- & \cite{li2020neural,li2020multipole,li2020fourier}\\
NNN & -- & \checkmark & \checkmark & -- & \checkmark & \cite{tao2018nonlocal}\\
\hline
NKN & \checkmark & \checkmark & \checkmark & \checkmark & \checkmark \\ \hline
\end{tabular}}
\end{center}
\caption{\small List of properties for GKNs, FNOs, NNNs, and NKNs.}
\label{tab:comparison}
\end{table}

\subsection{Problem statement: learning operators}

In this work, we aim to learn an operator between two functions, which can be seen as a mapping between two infinite dimensional spaces, given a collection of observed input-output function pairs. Let $D\subset\real^s$ be a bounded open set which is the domain of our input and output functions, we consider the problem of learning a general operator between two Banach spaces of functions taking values in $\real^{d_b}$ and $\real^{d_u}$, respectively. In what follows, we denote the input and output function spaces as $\mcB=\mcB(D;\real^{d_b})$ and $\mcU=\mcU(D;\real^{d_u})$, respectively. Let $\{\bb_j,\ub_j\}_{j=1}^N$ be a set of observations where the input $\{\bb_j\}\subset\mcB$ is a sequence of independent and identically distributed random fields from a known probability distribution $\mu$ on $\mcB$, and $G^\dag(\bb_j)=\ub_j(\xb)\in\mcU$, possibly noisy, is the output of the map $G^\dag:\mcB\to\mcU$. We aim to build an approximation of $G^\dag$ by constructing a nonlinear parametric map
$$
G(\cdot\,;\,\theta):\mcB\times\Theta\rightarrow\mcU,
$$
in the form of a NN, for some finite-dimensional parameter space $\Theta$. Here $\theta\in\Theta$ is the set of parameters in the network architecture to be inferred by solving the following minimization problem
\begin{equation}\label{eqn:opt}
\min_{\theta\in\Theta}\mathbb{E}_{\bb\sim\mu}[C(G(\bb;\theta),G^\dag(\bb))]\approx \min_{\theta\in\Theta}\sum_{j=1}^N[C(G(\bb_j;\theta),\ub_j)],
\end{equation}
where $C$ denotes a properly defined cost functional $C:\mcU\times\mcU\rightarrow\real$. Although $\bb_j$ and $\ub_j$ are (vector) functions defined on a continuum of points, with the purpose of doing numerical simulations, we assume that they are defined on a discretization of the domain $D$. In particular, for each data pair $(\bb_j,\ub_j)$ we assume observations of $\bb_j$ and $\ub_j$ are available on a $M-$point discretization of the domain defined as $D_j=\{\xb_1,\cdots,\xb_M\}\subset D$. With such a discretization, when learning governing laws, a popular choice the cost functional $C$ is the mean square error, i.e.,  the difference between $G(\bb_j;\theta)$ and $\ub_j$ in the $l^2$ norm defined on $D_j$. On the other hand, in image classification tasks, 
where $\bb_j$ represents the pixel values of the input image and $\ub_j$ the learnt feature function {which will be connected to a softmax layer for 
} classification, the cost functional (or classification loss) is usually the cross entropy loss \cite{haber2018learning}.

To stress the importance and challenges of learning operators, we now consider the problem of learning governing laws as an illustration. Let ${\rm L}_\bb$ be a differential operator depending on the parameter $\bb$ and consider the PDE 
\begin{equation}\label{eqn:pde}
\begin{aligned} 
-{\rm L}_\bb[\ub](\xb)=\fb(\xb),\quad&\xb\in D,\\
\ub(\xb)=0,\quad&\xb\in\partial D,
\end{aligned}
\end{equation}
for a given forcing term $\fb$. 
When the operator ${\rm L}$ is known, existing methods, ranging from the classical discretization of PDEs with known coefficients to modern ML approaches such as the basic version of physics-informed NNs \cite{raissi2019physics}, aim at finding the solution $\ub\in\mcU$ for a single instance of the parameter $\bb\in\mcB$. However, when the operator ${\rm L}$ is unknown, which is the case of interest here, the goal is to provide a {\it neural operator}, i.e. an approximated solution operator, $G(\cdot;\theta):\bb\rightarrow \ub$ that delivers solutions of the system for any input $\bb$. The latter problem not only is more realistic, as it is often the case that governing equations are not known for complex systems, but it is also a more challenging task for several reasons. First, in contrast to classical NN approaches where the solution operator is parameterized between finite-dimensional Euclidean spaces \cite{guo2016convolutional,zhu2018bayesian,adler2017solving,bhatnagar2019prediction,khoo2021solving}, neural operators are discretization and resolution independent. Therefore, \textit{no further modification or tuning will be required for different resolutions and discretizations} in order to achieve an equally accurate solution. Specifically, the neural operator generalizes to different grid geometries and discretizations. Second, for every new instance of $\bb$ neural operators requires only a forward pass of the network. Therefore, the optimization problem \eqref{eqn:opt} \textit{only needs to be solved once and the resulting NN can be utilized to solve for multiple instances of the input parameter}. This property is in contrast to the classical numerical PDE methods \cite{leveque2007finite,zienkiewicz1977finite,karniadakis2005spectral} 
and some ML approaches  \cite{raissi2019physics,weinan2018deep,bar2019unsupervised,smith2020eikonet,pan2020physics}, where the optimization problem needs to be solved for every new instance of the input parameter of a know differential operator ${\rm L}$. 
Lastly, of fundamental importance is the fact that neural operators can find solution maps regardless of the presence of an underlying PDE and only require the observed data pairs $\{(\bb_j,\ub_j)\}_{j=1}^N$. Examples include experimental measurements \cite{ranade2021generalized} and molecular dynamics simulations \cite{kim2019peri} for which an upscaled PDE is not available.


\subsection{Three relevant network architectures}

In this section, we discuss the network architecture of three baseline methods, namely, GKNs and the general integral kernel networks \cite{li2020neural,li2020multipole,li2020fourier}, NNNs \cite{tao2018nonlocal}, and multiscale CNNs \cite{haber2018learning}. To provide a consistent description of all three networks and illustrate their connections with the proposed NKN architecture, we describe each model following a formulation similar to the one presented in \cite{li2020neural,li2020multipole,li2020fourier}. First, we lift the input $\bb(\cdot)\in\mcB$ to a higher dimensional representation $\hb(\cdot,0)$ that corresponds to the first network layer; here, we identify the first argument of $\hb$ with space (the set of nodes) and the second argument with time (the set of layers). Second, we formulate the NN architecture in an iterative manner: $\hb(\cdot,0)\rightarrow \hb(\cdot,\Delta t)\rightarrow\hb(\cdot,2\Delta t)\rightarrow \cdots \rightarrow \hb(\cdot,T)$, where $\hb(\cdot,j\Delta t)$, $j=0,\cdots,L:=T/\Delta t$, is a sequence of functions representing the values of the architecture at each layer, taking values in $\real^{d}$. Third, the output $\ub(\cdot)\in\mcU$ is obtained by projecting $\hb(\cdot,T)$ onto $\mcU$. In what follows, we provide rigorous descriptions of these three steps.

Given an input vector field $\bb(\xb):\real^s\to\mbR^{d_b}$, we define the first network layer as 
$$\hb(\xb,0)=P(\xb,{\widetilde\bb}(\xb),\nabla{\widetilde\bb}(\xb))+\pb,$$
{where $\widetilde \bb$ represents a smoothed version of $\bb$, i.e. a continuous function of $\xb$. A common smoothing technique is given by Gaussian kernels \cite{li2020neural}. Note that this step would be helpful as inputs are usually in the form of vectors, e.g. function evaluations at grid points or pixel values of an image. As anticipated above, within each layer, we treat the nodes within a layer as a continuum so that we have an infinite number of nodes, i.e. a layer has infinite width. As such, each layer can be represented by a function of the continuum set of nodes \footnote{Considering an infinite width, i.e. defining neural networks in infinite-dimensional spaces, is not new and has been studied in, e.g., \cite{Williams1996,Roux2007}.} $D\subset\real^s$ . Then we denote the $l$-th network layer by $\hb(\xb,l\Delta t):\real^s\times \mathbb N^+\to\mbRd$, or, equivalently, $\hb(\xb,l\Delta t)=\hb(\xb,t):\real^s\times(0,T]\to\mbRd$. Here, $l=0$ (or equivalently, $t=0$) denotes the initial layer, whereas $t=L\Delta t$ (or $t=T$) denotes the last layer. The use of the symbol $t$ stems from the relationship that can be established between the network update and a time advancing scheme (or, in the limit of infinite layers, a dynamical system). 
The final output, computed using the network's last layer, is defined as $\ub(\xb)=Q\hb(\xb,T)+\qb$. Here, $P\in\real^{d\times(s+2d_b)}$, $Q\in\real^{d_u\times d}$, $\pb\in\real^{d}$ and $\qb\in\real^{d_u}$ are appropriately sized matrices and vectors that are part of the parameter set that we aim to learn}. We stress the fact that $\hb$ is a vector of dimension $d$ and, as such, a network layer has $d$ sets of nodes, each one associated with a component of $\hb$.

\paragraph{Graph kernel networks (GKNs)} Proposed in the context of learning governing equations, the GKN introduced in \cite{li2020neural} has foundation in the representation of the solution of a PDE by the Green's function. Here,  {for an $L-$layer NN,} the $l-$th layer network update is given by
\begin{equation}\label{eq:gkn}
\hb(\xb,l+1)=\sigma\left(R\hb(\xb,l)+\int_D k(\xb,\yb,\bb(\xb),\bb(\yb);\vb)\hb(\yb,l) d\yb + \mathbf{c}\right).
\end{equation}
Here, $\sigma$ is an activation function, $R\in\real^{d\times d}$ is a tunable tensor, $\cb\in\real^d$ a constant vector and $k\in\real^{d\times d}$ a tensor kernel function that takes the form of a (usually shallow) NN whose parameters $\vb$ are to be learned. In GKNs, different layers share the same parameters $\vb$, $R$ and $\cb$, and the kernel $k$ is therefore layer-independent. This network update resembles the original ResNet block \cite{He2016Resnet} where the usual discrete affine transformation is substituted by a continuous integral operator. Differently from the networks that we consider later on, unless $\sigma$ is the identity operator, we cannot establish a connection between \eqref{eq:gkn} and a discretized PDE or an ordinary differential equation. {While in the original version of GKNs the integral is extended to the whole set $D$, for efficiency purposes, restrictions to a ball of radius $r$ centered at $\xb$, i.e. $B_r(\xb)$, can also be considered, keeping in mind that this choice might compromise the accuracy}.

Single-layer and shallow GKNs have been shown to be successful in learning governing equations for, e.g., the Darcy \cite{li2020neural} and Burger \cite{li2020multipole} equations. The most notable advantage of this approach is that the learnt network parameters are resolution-independent: the learned $R$, $\cb$, and $\vb$ are optimal even when used with different resolutions, i.e. with different partitions/discretizations of the feature space $D$. Even though not exploited in \cite{li2020neural}, resolution-independence can be critical in image transfer learning tasks. However, in the presence of complex learning tasks, shallow networks might not be sufficiently accurate, so that deep networks become mandatory. As we illustrate in numerical studies of Section \ref{sec:experiments} the major drawback of GKNs is their instability with respect to increasing number of layers; in fact, as the GKN becomes deeper, either there is no gain in accuracy or increasing values of the loss function occur. 


\paragraph{Fourier neural operators (FNOs)} We mention in this paragraph also a new variant of integral neural operators, namely the Fourier neural operator (FNO) proposed in \cite{li2020fourier}, where the integral kernel $k$ is parameterized in Fourier space. In particular, FNO drops the dependence of kernel $k$ on the input $\bb$ and assumes that $k(\xb,\yb;\vb):=k(\xb-\yb;\vb)$. The integral operator in \eqref{eq:gkn} then becomes a convolution operator so that $k$ can be parameterized in Fourier space. The corresponding $l-$th layer update is then given by
\begin{equation}\label{eq:fno}
\hb(\xb,l+1)=\sigma\left(R(l)\hb(\xb,l)+\mathcal{F}^{-1}(\mathcal{F}(k(\cdot;\vb_l))\cdot \mathcal{F}(\hb(\cdot,l)))(\xb)+ \mathbf{c}(l)\right),
\end{equation}
where $\mathcal{F}$ and $\mathcal{F}^{-1}$ denote the Fourier transform and its inverse, respectively. {Here we use $R(l)$, $\mathbf{c}(l)$ and $\vb_l$ to highlight the fact that in FNOs, each layer has different parameters (i.e. different kernels, weights and biases).} This is in contrast with the layer-independent kernel in the original GKNs. As a consequence when the number of layers increases, the memory consumption of FNOs increases, which makes the training process of FNOs more challenging and potentially prone to over-fitting.

\paragraph{Nonlocal neural networks (NNNs)} 
With the purpose of circumventing the instability properties of a nonlocal network architecture proposed in \cite{Wang2018nonlocal} (similar to a GNN), the paper \cite{tao2018nonlocal} proposes a modified nonlocal network architecture where the nonlocal operator is augmented in such a way that it corresponds to a discrete nonlocal diffusion operator. Here, the set of nodes is not treated as a continuum and the {\it discrete} network update is defined as
\begin{equation}\label{eq:nnn}
\hb_i(l+1)=\hb_i(l)+ R(l)\sum_{j=1}^M k(i,j) (\hb_j(l)-\hb_i(l)),
\end{equation}
where the subscript $i$ indicates the node and $l=0,\cdots,L$ still denotes the layer and where the only parameters to be learned are the entries of the ``weight'' matrix $R(l)\in \real^{d\times d}$ at every layer. The pairwise affinity function $k(i,j)\in\real$ is given and is usually a symmetric, nonnegative function. The introduction of the term $(\hb_j(l)-\hb_i(l))$ in \cite{tao2018nonlocal}, in place of $\hb_j(l)$ only as in \cite{Wang2018nonlocal}, significantly improves the accuracy of the network when utilized for image processing tasks. In particular, the network update \eqref{eq:nnn}, also called ``nonlocal block'', is used within more standard networks, such as ResNets, with the purpose of improving their accuracy thanks to the fact that nonlocal blocks take into account long-range node interactions. The major drawback of this architecture, being formulated at the discrete level, is that it cannot be resolution independent and, hence, the learned parameters are not optimal when utilized within networks of different width. As this property can only be achieved in the presence of continuous operators, we report for the sake of completeness the continuous version of the NNN in \eqref{eq:nnn}
\begin{equation}\label{eq:nnn-cont}
\hb(\xb,l+1)=\hb(\xb,l)+R(l)\int_D k(\xb,\yb)(\hb(\yb,l)-\hb(\xb,l)) d\yb,
\end{equation}
where $k$ is a given, symmetric, nonnegative function of its arguments. While a comparison of \eqref{eq:nnn-cont} with GKNs has not been conducted in the literature, we expect the latter to be outperformed in the limit of deep networks for stability reasons. However, in the shallow case, GKNs are likely to perform better due to their increased descriptive power as the kernel $k$ is part of the unknowns while in \eqref{eq:nnn-cont} it is given.

We point out that the GKN architecture \eqref{eq:gkn} can also be seen as the continuous version of the nonlocal network proposed in \cite{Wang2018nonlocal}, where the authors introduce a discrete update based on convolution operators acting on nodes, at the discrete level. As such, the approach in \cite{Wang2018nonlocal} not only does not feature resolution independence, but also shows instabilities in the deep network limit, as pointed out in \cite{tao2018nonlocal}.

\paragraph{Multiscale CNN} 
Paper \cite{haber2018learning} introduces a new approach to training CNNs that allows for       ``learning across scales'' (i.e. for independence with respect to width and depth). By reinterpreting the CNN architecture as a discretization of a time-dependent nonlinear differential equation, the network depth corresponds to advancing in time. When the network is stable, the idea of \cite{haber2018learning} is to interpolate and reuse optimal parameters of a shallow network into a deeper one. More specifically, by identifying the number of layers with the number of time steps in a time-discretization scheme, they employ multilevel learning algorithms that accelerate the training of deep NNs by solving a series of learning problems from shallow to deep architectures. We refer to the resulting technique as shallow-to-deep learning. Formally, let $t=l\Delta t$; then, the $(l+1)$-th network layer is given by 
\begin{equation}\label{eq:cnn}
\hb(t+\Delta t)=\hb(t)+ \Delta t\, \sigma(R(\kb;t)\hb(t)+\cb),
\end{equation}
where $\sigma$ is an activation function, $R(\kb;t)\in\real^{d\times d}$ is a convolution matrix (a circulant matrix that depends on the convolution kernel $\kb$), and $\cb$ is a bias vector. It is easy to see that by diving both sides of \eqref{eq:cnn} by $\Delta t$, the term $(\hb(t+\Delta t)-\hb(t))/\Delta t$ corresponds to the discretization of a first order derivative so that this architecture can indeed be interpreted as a nonlinear differential equation in the limit of deep layers, i.e. as $\Delta t\to 0$. Thus, when the real parts of the eigenvalues of the convolution and the time steps are sufficiently small, this architecture is stable with respect to the number of layers. The shallow-to-deep learning mentioned above corresponds to training the network for increasing values of network layers and using optimal parameters obtained {with $L$ layers as initial guesses for the $\tilde{L}$-layer CNN 
}, after appropriate scaling and interpolation across layers. {Here $\tilde{L}>L$.}

{We point out that, even though successful in image processing tasks, standard CNNs are not resolution independent unless appropriately modified (via, e.g., multiscale or multigrid methods \cite{haber2018learning}). Furthermore, due to the fact that interactions between nodes occur only in limited node-windows, they are not as flexible as, e.g., NNNs where node-to-note interactions are extended to the {whole node set}.}

\section{Nonlocal Kernel Networks (NKN)}\label{sec:nkn}
To overcome the limitations of the architectures mentioned in Section \ref{sec:background} and still preserve their benefits, in this section, we propose a new, stable, and resolution-independent network update. We first describe the Nonlocal Kernel Network (NKN) architecture and review relevant definitions and results of the nonlocal vector calculus. These tools are then used to prove the stability properties of the proposed network architecture in the deep-layer limit. Lastly, we illustrate how to perform shallow-to-deep training, exploiting the stability of the network in the limit of deep layers.

\subsection{The network architecture}

Using the same notation of Section \ref{sec:background}, we introduce the network update for the proposed NKN architecture. Let $t=l\Delta t$, being $l$ the current layer, and, as before, let $\xb\in D$ span the continuum set of nodes within each layer. We propose the following iterative network update formulation
{\begin{equation}\label{eq:NKN}
\hb(\xb,t+\Delta t)=\hb(\xb,t)+ {\Delta t}\left(\int_D k(\xb,\yb,\bb(\xb),\bb(\yb);\vb)(\hb(\yb,t)-\hb(\xb,t)) d\yb-R(\xb;\wb)\hb(\xb,t)+\cb\right).
\end{equation}}
As for GKNs, the kernel tensor function $k\in\real^{d\times d}$ is modeled by a NN parameterized by $\vb$. To enhance the descriptive power and stability properties of the network a reaction term is added on the right-hand side. Here, the tensor function $R\in\real^{d\times d}$ is modeled by another NN parameterized by $\wb$. Both $k$ and $R$ are usually shallow NNs, such as the multilayer perceptron (MLP) employed in our numerical examples. Their depth and width depend on the specific application and will be specified later on. Note that the integral operator on the right-hand side of \eqref{eq:NKN} can be interpreted as a nonlocal Laplacian $\mathcal L_k[\cdot]$, as clarified in the following section. The NKN architecture above preserves the continuous, integral treatment of the interactions between nodes that characterizes GKNs and replaces the integral operator acting on $\hb(\yb,t)$ in that formulation with the continuous version of the nonlocal diffusion operator introduced in \cite{tao2018nonlocal}, as defined in \eqref{eq:nnn-cont}. While the resemblance with GKNs enables resolution independence with respect to the inputs, the use of the nonlocal operator provides rigorous analysis tools that will allow us to show that the architecture is stable in the deep network limit. 
We point out that in our formulation the network parameters are not time-dependent, i.e. they are constant across the layers; this feature enables the straightforward application of the shallow-to-deep initialization technique and reduces the computational effort and memory allocation. The idea using constant parameters across layers was also proposed in implicit networks \cite{el2021implicit,bai2019deep,winston2020monotone,bai2020multiscale} where fixed-point methods are employed as an efficient training procedure.

In Table \ref{tab:comparison} we summarize relevant properties of NKNs in comparison with GKNs and NNNs. These statements are confirmed and illustrated by both the theoretical results presented in the following sections and by the numerical tests reported in Section \ref{sec:experiments}. In summary, being resolution independent and stable in the limit of deep layers makes the NKN's architecture a viable tool for both PDE learning and image processing tasks.

\subsection{Connection to the nonlocal vector calculus}\label{sec:nonlocal-calculus}

In this section we recall important concepts of the nonlocal vector calculus that are useful to prove stability properties of the proposed network architecture \eqref{eq:NKN}. Note that, for the sake of simplicity, we limit our description to the scalar case for which $h:\real^s\times (0,T]\to \mbR$, {although the description and analysis can be extended to the vector case $\hb:\real^s\times (0,T]\to\real^d$.} For more details on this topic we refer the reader to the review articles \cite{DeliaDuEtAl2020_NumericalMethodsNonlocalFractionalModels,du2013nonlocal}.

\begin{figure}[t]
	\centering
	\includegraphics[width=0.3\columnwidth]{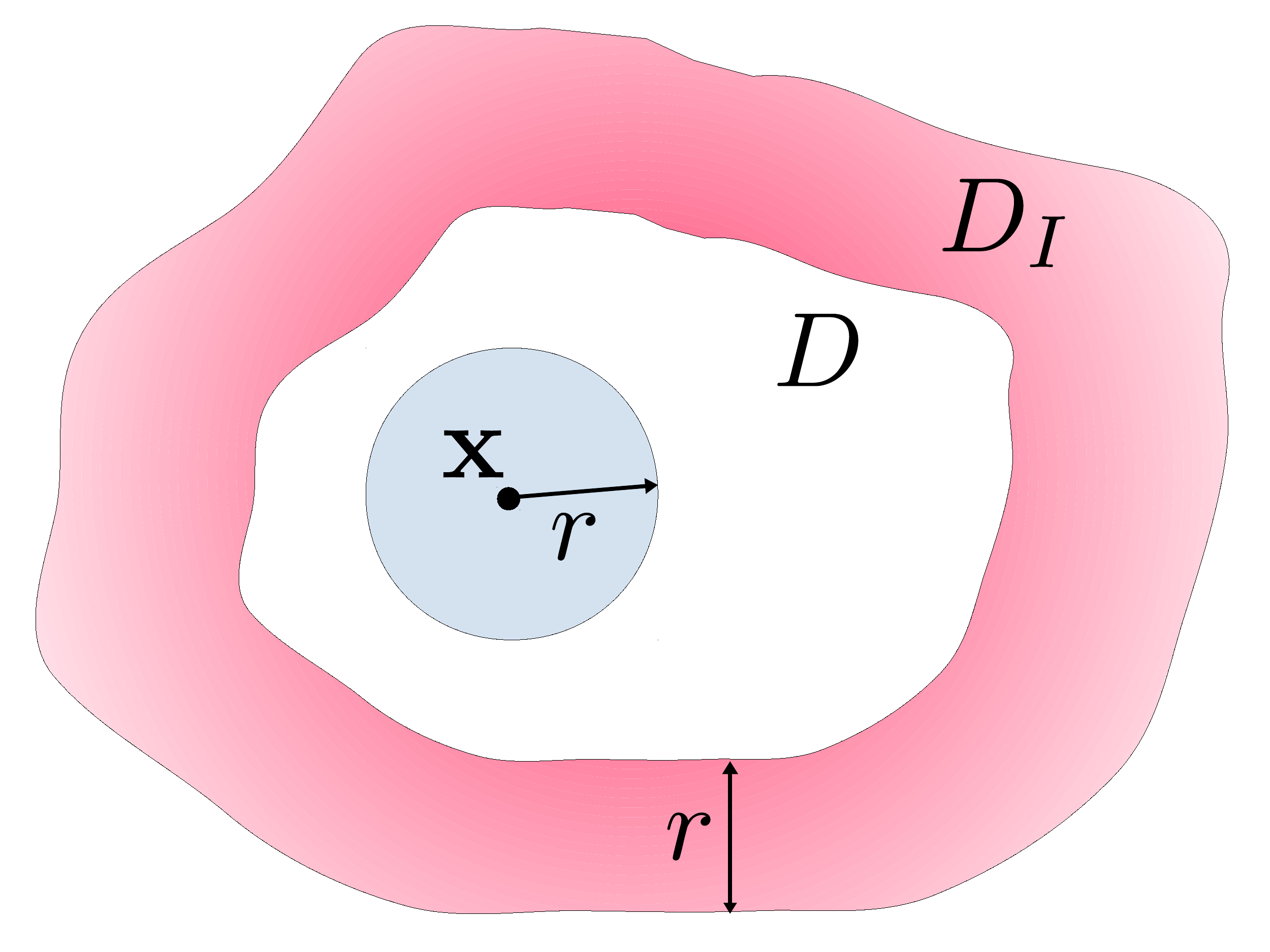}
	\caption{Two dimensional $(s=2)$ illustration of domain $D$, interaction domain $D_I$, and nonlocal neighborhood $B_r(\xb)$.}
\label{fig:domain}
\end{figure}

The main feature of nonlocal models is that every point in a domain of interest, $D\in\real^s$, interacts with a {\it nonlocal neighborhood} of points, usually described by the Euclidean ball $B_r(\xb)$. As a consequence, when solving a nonlocal equation in a bounded domain, boundary conditions must be prescribed on a {\it nonlocal boundary}, that accounts for all the points outside of $D$ that interact with $D$. We refer to this set of points as interaction domain and denote it by $D_I$. When the nonlocal neighborhood is $B_r(\xb)$, the interaction domain corresponds to a layer of thickness $r$ surrounding the domain (see Figure \ref{fig:domain}), where nonlocal boundary conditions must be prescribed to guarantee well-posedness of solutions. We denote the union of domain and interaction domain by $\overline D$.

The nonlocal vector calculus \cite{d2021towards,Delia2017,Du2012}, provides a variational settings that allows one to study nonlocal equations in a very similar way as the classical PDEs are analyzed. Given a square integrable kernel function $k:\overline D\times\overline D\to\mathbb R^+$ with compact support in $B_r(\xb)$, the nonlocal Laplacian operator is defined as 
\begin{equation}\label{eq:nonlocal-Laplacian}
\mcL_k[h](\xb) = \int_{\overline D} k(\xb,\yb)(h(\yb,t)-h(\xb,t)) d\yb.
\end{equation}
In this work we consider parabolic nonlocal diffusion-reaction equations due to the resemblance of our network architecture with such equations in the limit of deep layers. We define the strong form of such an equation as follows: given a reaction term $R:D\to\mbR$, such that $0<R_0\leq R(\xb) \leq R_1<\infty$, a constant forcing term $c\in\mbR$, a kernel $k$ with the above properties and an initial state $h_0(\xb)$, find $h:D\to\mbR$ such that
\begin{equation}\label{eq:nonlocal-parabolic_1}
\begin{aligned}
&\dfrac{\partial h}{\partial t}(\xb,t) -\mcL_k[h](\xb,t) +R(\xb)h(\xb,t)=c, & (\xb,t)\in D\times[0,T], \\[2mm]
&h(\xb,0) = h_0(\xb), & \xb\in \overline D, \\[2mm]
&h(\xb,t) = 0, & (\xb,t)\in D_I\times[0,T],
\end{aligned}
\end{equation}
where the last condition is the nonlocal counterpart of a homogeneous Dirichlet boundary condition, prescribed on the interaction domain $D_I$. 

We denote by $\mcA$ the nonlocal elliptic operator $\mcA_k[\cdot]= -\mcL_k[\cdot]+R(\xb) \, [\cdot]$ that features a nonlocal diffusion component and a (classical) reaction component, respectively. By using the nonlocal vector calculus we can analyze the variational form of \eqref{eq:nonlocal-parabolic_1}, that we introduce next. Given a kernel $k$ defined as above, a reaction coefficient $R\in L^\infty(D)$ such that $0<R_0\leq R(\xb) \leq R_1<\infty$, a constant $c\in\mbR$, and an initial state $h_0\in L_0^2(\overline D)$, the weak solution $h\in L^2(0,T;L^2_0(\overline D))$ of \eqref{eq:nonlocal-parabolic_1} satisfies, for all $\eta\in L^2_0(\overline D)$
\begin{equation}\label{eq:parabolic-weak_1}
\int_D \dfrac{\partial h}{\partial t}\eta \,d\xb +
\int_D \mcA[h]\eta\,d\xb = \int_D c \,\eta \,d\xb,
\end{equation}
where $L^2_0(\overline D)$ is the space of square integrable functions on $\overline D$ that are zero on $D_I$. By using nonlocal integration by parts \cite{Du2012}, we have that
$$
\int_D \mcL_k[h]\eta\,d\xb = \iint_{\overline{D}\times \overline{D}}(h(\yb,t)-h(\xb,t))(\eta(\yb,t)-\eta(\xb,t))k(\xb,\yb)\,d\yb\,d\xb:= a_k(h,\eta),
$$
where we have exploited the fact that $h=0$ in $D_I$.

The nonlocal vector calculus theory \cite{Du2012} guarantees that for square integrable, compactly supported, kernel functions, the bilinear form $a_k(\cdot,\cdot)$ induces an inner product in $L^2{(\overline{D})}$, or, in other words, there exist positive constants $\underline{C}$ and $\overline{C}$ such that
\begin{equation}\label{eq:norm-equivalence}
\underline{C} \|\eta\|_{L^2(\overline{D})}\leq \sqrt{a_k(\eta,\eta)}\leq\overline{C}\|\eta\|_{L^2(\overline{D})},
\quad \forall\,\eta\in L^2_0(\overline D).
\end{equation}
This property implies that the bilinear form associated with $\mcA$ is coercive and continuous in the $L^2$ metric, yielding well-posedness of equation \eqref{eq:parabolic-weak_1}.

\subsection{NKNs as stable parabolic nonlocal equations}\label{sec:stability}
In this section we analyze the mathematical properties of the NKN model. Without loss of generality, and to be consistent with Section \ref{sec:nonlocal-calculus}, we consider the case for which $\hb:D\to\mbR$. Thus, we denote the network by $h$. With the purpose of highlighting the connection to a time discretization scheme, we divide both sides of \eqref{eq:NKN} by $\Delta t$ and rewrite the NKN update as 
\begin{equation}\label{eq:euler}
\dfrac{h(\xb,t+\Delta t)-h(\xb,t)}{\Delta t} -\mcL_k[h](\xb,t) +R(\xb)h(\xb,t)=c.
\end{equation}
Here, we note that the first term on the left-hand side corresponds to the explicit Euler discretization of a time derivative. As such, we can claim that the limit as $\Delta t\to 0$ of \eqref{eq:euler} corresponds to
\begin{equation}\label{eq:nonlocal-parabolic}
\dfrac{\partial h}{\partial t}(\xb,t) -\mcL_k[h](\xb,t) +R(\xb)h(\xb,t)=c.
\end{equation}
As described in Section \ref{sec:nonlocal-calculus}, \eqref{eq:nonlocal-parabolic} is a parabolic nonlocal equation with nonlocal elliptic operator $\mcA_k[\cdot]= -\mcL_k[\cdot]+R(\xb) \, [\cdot]$.  Standard variational theory and the nonlocal vector calculus enable the analysis of the weak form of  \eqref{eq:nonlocal-parabolic} for which we prove well-posedness and a-priori bounds on the solution in the following theorem.

\begin{thm}\label{thm}
{Let $k\in L^2(\overline D\times\overline D)$, $R\in L^\infty(D)$ such that $0<R_0\leq R(\xb) \leq R_1<\infty$, $c\in\mbR$, and $h_0\in L_0^2(\overline D)$. Then, problem \eqref{eq:nonlocal-parabolic_1} is well-posed and, in particular, for all $t>0$,}
\begin{equation}\label{eq:a-priori_1}
\|h(\cdot,t)\|^2_{L^2(\overline{D})}+ \widetilde C\int_0^t \|h(\cdot,s)\|^2_{L^2(\overline{D})}\,ds  \leq \|h_0\|^2_{L^2(\overline{D})} +\dfrac{c^2|\overline{D}|t}{2\widetilde C},
\end{equation}
{\it where $\widetilde C=\underline{C}^2(\underline{C}^2+R_0)$.}
\end{thm}
\begin{proof}
Property \eqref{eq:norm-equivalence} and the bounds on the reaction term $R$ imply that the bilinear form associated with the operator $\mcA$ is coercive and continuous in $L^2_0(\overline D)$. In fact, the following inequalities hold
\begin{equation}\label{eq:coer-cont}
\begin{aligned}
&\int_D \mcA[h]h\,d\xb \geq (\underline{C}^2+R_0)\|h\|^2_{L^2(\overline{D})}
& \quad \text{(coercivity),}\\
&\left| \int_D \mcA[h]\eta\,d\xb \right| \leq
(\overline{C}^2+R_1)\|h\|_{L^2(\overline{D})}
\|\eta\|_{L^2(\overline{D})} &\quad  \text{(continuity).}
\end{aligned}
\end{equation}
Continuity and coercivity, combined with the continuity of the functional $\int_D c\eta d\xb$, are sufficient conditions for the well-posedness of equation \eqref{eq:parabolic-weak_1}. Furthermore, by using standard arguments of variational PDE theory (see, e.g., \cite{Delia2017,d2021analysis}), paper \cite{Delia2017} shows that the unique solution $h\in L^2(0,T;L^2_0(\overline D))$ satisfies the a priori bound \eqref{eq:a-priori_1} for all $t>0$. We note that the theory developed in \cite{Mengesha2013} allows us to extend this result to sign-changing kernels, like the one utilized in this work. Finally we point out that the arguments used in this proof can be extended to the vector case $\hb\in\real^d$.
\end{proof}

As a consequence, for any given final time, the solution $h(\xb,t)$ is guaranteed to be bounded. This fact proves the stability of the NKN model; the latter will be confirmed by our numerical experiments in Table \ref{tab:eigen} of Section \ref{section:pde}. 

\subsection{Shallow-to-deep NKN learning}

The stability properties of NKNs allow us to consider deep networks and to exploit efficient initialization techniques such as the shallow-to-deep approach introduced in Section \ref{sec:background}. Let $R_L\in\real^{d\times d}$, $\cb_L\in \real^d$ and $k_L(\xb,\yb,\bb(\xb),\bb(\yb))$ be the optimal network parameters obtained by training a NKN of depth $L$. With the purpose of improving the accuracy of the network, we increase the number of layers (or equivalently, time steps), and train a new network of depth $\widetilde{L}>L$. The idea of the shallow-to-deep technique is to interpolate in time (or across layers) the optimal parameters obtained at depth $L$ and to scale them in such a way that the final time of the differential equation remains unchanged. In our specific setting, due to the fact that the network parameters are not time dependent, this technique simply corresponds to initializing the (deeper) $\widetilde L$-layer network by $R_L$, $\cb_L$, and $k_L$.

\section{Numerical experiments}\label{sec:experiments}

In this section, we illustrate the superior performance of NKNs in both learning governing laws and image classification tasks, and compare it to baseline approaches. Our numerical experiments are performed on a machine with 2.8 GHZ 8-core CPU and a single Nvidia V100 GPU.

\subsection{Learning governing laws}\label{section:pde}

To demonstrate the stability of NKNs in the deep layer limit and its superiority with respect to other methods, we consider two learning examples employed in \cite{li2020neural} for GKNs, and compare the performance of NKNs with GKNs and FNOs for layers from $L=1$ to $L=32$. Specifically, we consider the problem of learning neural operators that act as solution maps for the PDE \eqref{eqn:pde}, without any prior knowledge on the PDE itself, but solely on the basis of an input-output data set. The training set consists of $N$ pairs of input parameter functions and solutions $\{\bb_j(\xb_i),\ub_j(\xb_i)\}_{j=1}^N$ available at $\xb_i\in D_j:=\{\xb_i,i=1,\cdots,M\}\subset D.$ For simplicity and without loss of generality, we focus on the simple setting where all function pairs are evaluated on the same, structured grid of points with grid size $\Delta x$, and we refer to is as $D_{\Delta x}=D_j$ for all $j=1,\cdots,N$.
We recall that our major goal is to design a network architecture that is stable in the limit of deep layers and resolution independent, so that we can reach increasingly better levels of accuracy for deeper networks and predict an equally accurate solution $\ub$ when using different values of $\Delta x$.

For the implementation of GKNs and NKNs, we use the pytorch library provided in \cite{li2020neural}. For FNOs, we use the Pytorch package provided in \cite{li2020fourier}. The optimization is performed with the Adam optimizer. To conduct a fair comparison, for each method, we have tuned the hyperparameters, including the learning rates, the decay rates and the regularization parameters, to minimize the training loss. Furthermore, for each example and each method we repeat the numerical experiment for 5 different random initializations, and report the averaged relative mean squared errors and their standard error. With the purpose of having a compact presentation of the results, we report the errors in plots, as functions of the number of NN's layers. A more detailed error comparison is provided in Tables \ref{tab:1DPoisson_new}-\ref{tab:2DDarcy_reso_more} of \ref{sec:newapp_pde}.

\subsubsection{Example 1: 1D Poisson's equation}

\begin{table}[]
    \centering
    \begin{tabular}{|c|c|c|c|c|c|c|}
    \hline
    Model & $L=1$ &$L=2$&$L=4$&$L=8$&$L=16$&$L=32$ \\
    \hline
    GKN  & 66.82k & 66.82k & 66.82k & 66.82k & 66.82k & 66.82k \\
    NKN & 67.02k & 67.02k & 67.02k & 67.02k & 67.02k & 67.02k\\
    FNO &  78.33k & 148.03k & 287.43k & 566.21k & 1.12M & 2.24M\\
    \hline
    \end{tabular}
    \caption{Example 1: 1D Poisson's equation. Number of trainable parameters for each model.}
    \label{tab:1d_param}
\end{table}

\begin{table}[]
    \centering
    \begin{tabular}{|c|cc|cc|}
    \hline
    \multirow{2}{*}{Depth} & \multicolumn{2}{c|}{NKN}
    & \multicolumn{2}{c|}{GKN}\\
    \cline{2-5}
    & max eigenvalue & min eigenvalue & max eigenvalue & min eigenvalue\\
    \hline
    1 & 1.6012 & -0.1183 & 0.9958 & -6.3943\\
    2 & 2.2862 & 0.8871 &  5.6733 & 1.7302\\
    4 & 3.7286 & 1.6576& 13.3126& -0.2470\\
    8 & 5.3263 & 2.1790& 27.1865 & -3.6409\\
    16 & 6.6001 & 2.4899& 51.3217 & -0.2268\\
    32 & 7.3043 & 2.5560& 48.8081 & 36.3245\\
    \hline 
    \end{tabular}
    \caption{Example 1: 1D Poisson's equation. Maximum and minimum eigenvalues of the (linearized) amplification operators for NKNs and GKNs, from the $l-$th layer to the $(l+1)-$th layer on an original training sample.}
    \label{tab:eigen}
\end{table}

In this example we consider ${\rm L}=\frac{\partial^2}{\partial x^2}$, i.e. the one-dimensional Poisson's equation, in $D=[0,1]$ taking the form:
\begin{equation}\label{eqn:1dpoisson}
\begin{aligned} 
-\frac{\partial^2 u}{\partial x^2}(x)=f(x),\quad&x\in D,\\
u(x)=0,\quad&x\in\partial D.
\end{aligned}
\end{equation}
We aim to learn the operator mapping the loading function $f(x)$ to the solution $u(x)$. 
The training data set consist of $N=500$ pairs of $f_j(x_i)$ and $u_j(x_i)$ for $x_i\in D_{\Delta x}$, where $D_{\Delta x}=\{0.01i|i=0,\cdots,100\}$, a set of 101 uniformly distributed points in $D$. To generate each sample pair $\{f_j(x),u_j(x)\}$, we first set
$$
u_j(x) = \sum_{k=0}^{100} \widehat{u}_{k,j} \cos(2\pi kx)
$$
with $\widehat{u}_{k,j}$ being constant coefficients. For each $k \in \{1,\cdots,100\}$, $\widehat{u}_{k,j}$ is randomly generated as $\widehat{u}_{k,j} \sim \mathcal{U}[0,\exp(-0.1 k^2)]$, the uniform distribution on $[0,\exp(-0.1 k^2)]$. The term $\widehat{u}_{0,j}$ is chosen such that the boundary condition $u_j(0)=u_j(1)=0$ is satisfied. Then $f_j(x)$ is obtained from $u_j(x)$ via a numerical Fourier transform and sample pairs are obtained by evaluating $u_j$ and $f_j$ at points on $D_{\Delta x}$. To validate the performance of the trained model, we generate $100$ additional pairs following the same procedure used for the training set. Note that the solution of \eqref{eqn:1dpoisson} can be represented as
\begin{equation}
    u(x)=\int_D G_b(x,y)f(y)dy
\end{equation}
where $G_b(x,y):=\frac{1}{2}(x+y-|y-x|)-xy$ is the Green's function. The integral form above suggests that a 1-layer NKN can provide an exact solution map by setting the dimension of $\hb$ as $d=1$, the initial layer and the final output as
$$h(x,0)=f(x),\quad u(x)=h(x,1),\quad T=1,\quad L=1,$$
and the network update formulation as
$$
c=0,\quad 
R(x)=1-\int_D G_b(x,y)dy,\quad 
{\rm and} \quad 
k(x,y,f(x),f(y))=G_b(x,y).
$$
Therefore, in principle, when the number of training pairs $N\rightarrow\infty$ and the integral on $D$ is evaluated exactly, a 1-layer NKN can provide an exact map. 
Note that, for different choices of parameters, this statement holds true for GKNs as well. It is important to stress that for both networks increasing the number of layers {would not yield significant improvements in the prediction accuracy 
}; instead, in general, it may generate instabilities that might compromise the network performance. This fact makes the 1D Poisson equation the best candidate example to explore the network stability when the number of NN layer increases. Note that these considerations do not apply to more complex learning examples such as the prediction of solutions in highly heterogeneous environments, where, deeper and deeper networks are required for accuracy purposes.


Inspired by the discussion above and following \cite{li2020neural}, we set $d=1$, $T=1$ and initialize our network by 
$
h(x,0)=P(x,f(x))+p.
$
Since the ground-truth kernel (the Green's function $G_b$) is independent of $f$, we set the kernel $k(x,y,f(x),f(y)):=k(x,y)$. By setting $\Delta t=1/L$, the NKN network update reads
$$h(x,t+\Delta t)=h(x,t)+\Delta t\left(-R(x;\wb)h(x,t)+\int_D k(x,y;\vb)(h(y,t)-h(x,t)) dy + c\right).$$
The inner kernel network $k(x,y):\real^{2}\rightarrow \real$ is parameterized as a 3-layer feed forward network with widths $(2,256,256,1)$ and ReLU activation. The reaction network $R(x):\real\rightarrow\real$ is taken as a 2-layer feed forward network with widths $(1,64,1)$ and ReLU activation. The solution $u$ is then computed as $u(x)=Q(h(x,L))+q$. Here $P$, $p$, $Q$, $q$ and $c$ are all trained. We apply the shallow-to-deep training technique to initialize the optimization problem when the number of layers $L>1$. Specifically, we start from depth $L = 1$, train until the loss function reaches a plateau and use the estimated parameters to initialize the parameters for $L=2$, until $L=32$ (recall that the optimal parameters do not depend on the layer/time).

\begin{figure}
    \centering
    \includegraphics[width=.48\textwidth]{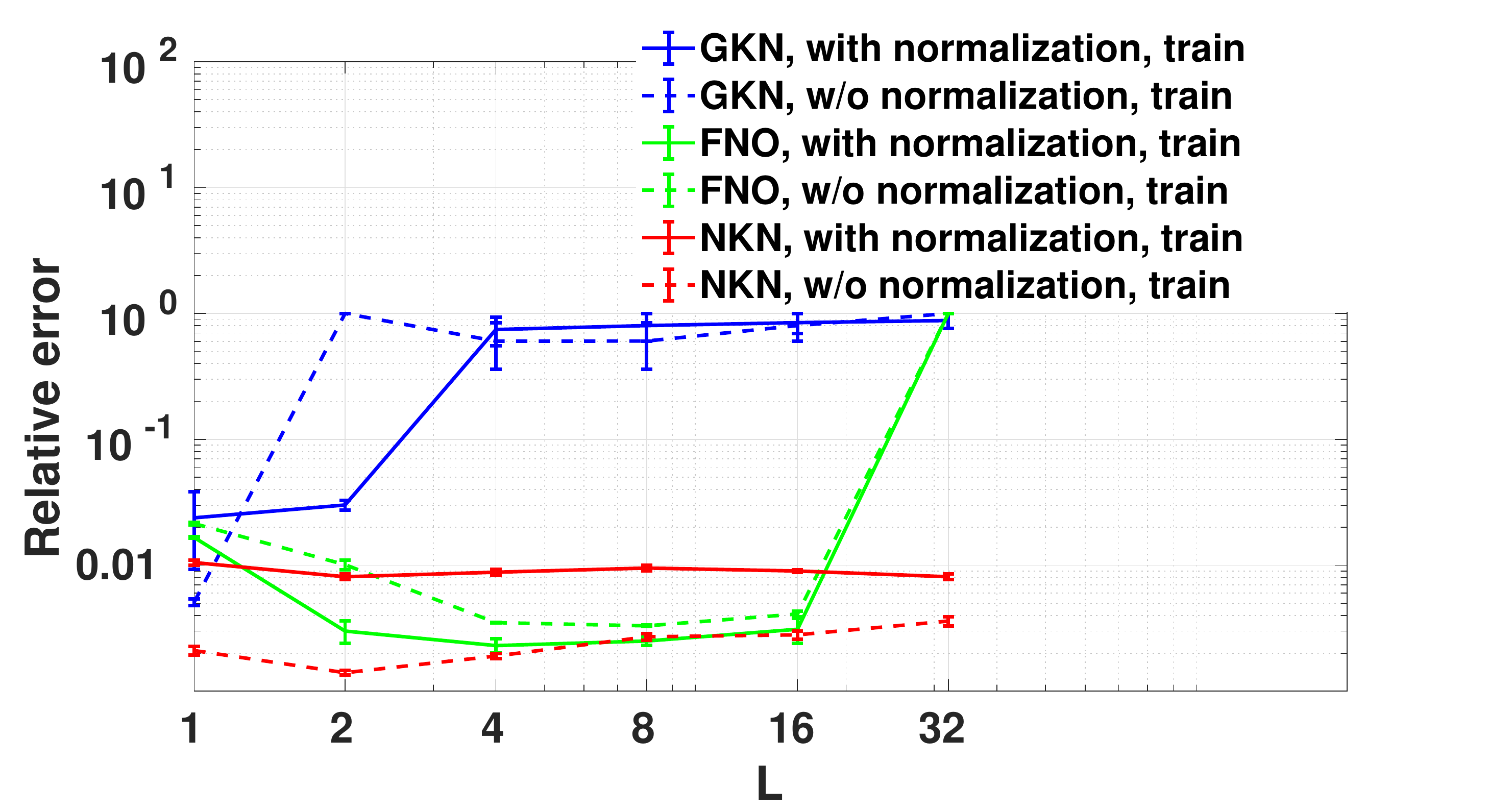}
    \includegraphics[width=.48\textwidth]{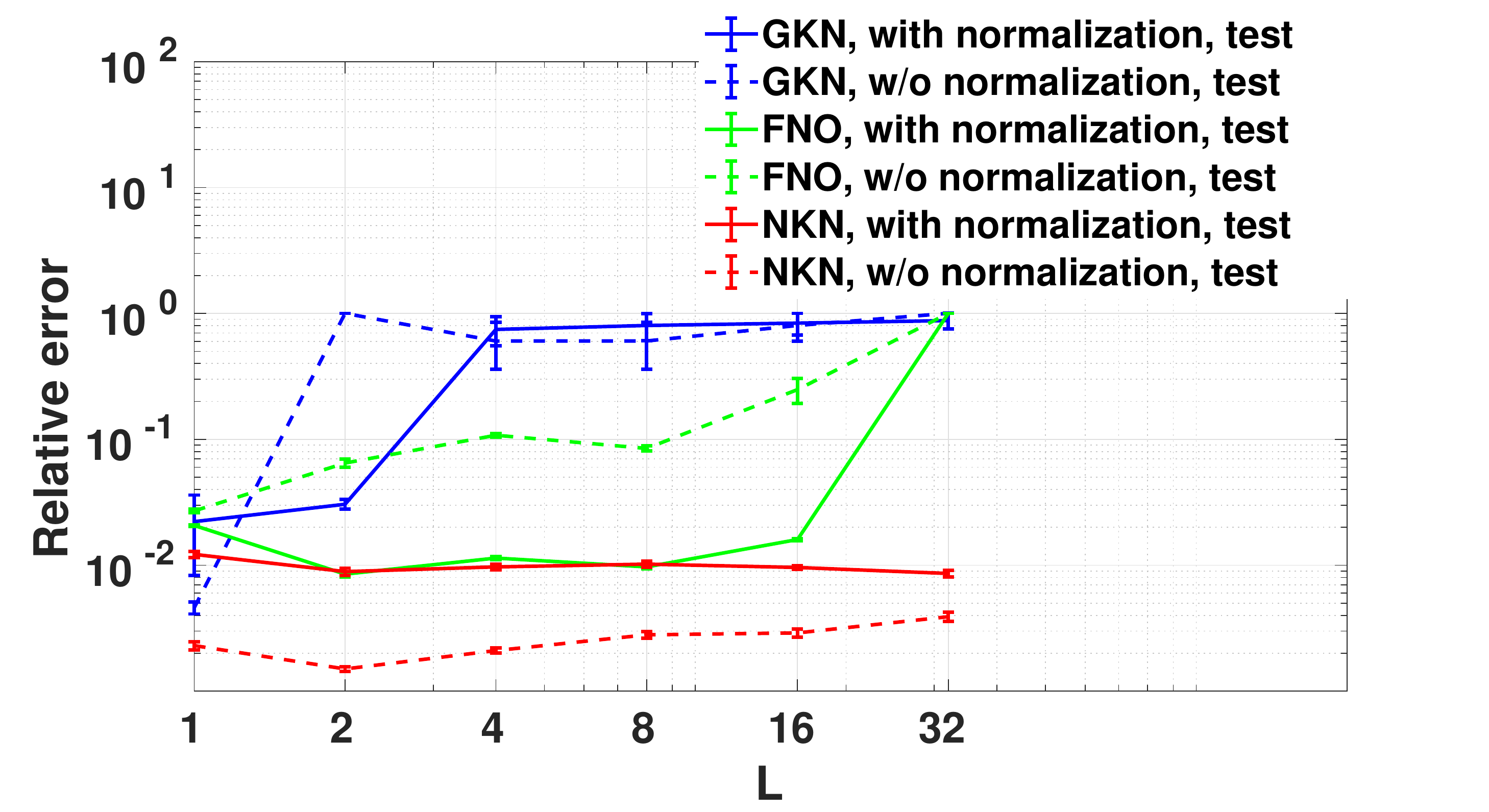}
    \caption{Example 1: 1D Poisson's equation. Comparison of relative mean squared errors from GKNs, FNOs, and NKNs. Error bars represent standard errors over $5$ simulations. Left: errors on training dataset. Right: errors on test dataset.}
    \label{fig:poisson_loss}
\end{figure}

To investigate the stability properties of each neural operator learning models, we compare the performance of NKNs with GKNs and FNOs as the number of layers increases. For all methods we train until the loss function reaches a plateau ({10000} epochs at most). In Figure \ref{fig:poisson_loss} we present the averaged relative mean squared errors for each model as a function of $L$; the number of trainable parameters is provided in Table \ref{tab:1d_param}. To study the impact of normalization, we report learning results on both normalized (denoted as the ``with normalization'' cases) and original (denoted as the ``w/o normalization'' cases) training data sets.

\textit{Comparison between GKNs and NKNs.} In the left plot of Figure \ref{fig:poisson_loss} we compare the training errors, from which we can observe the very poor performance of GKNs for $L>2$. Note that reducing the learning rate or increasing the learning epochs does not mitigate this convergence issue. In contrast, for increasing values of $L$, NKNs are stable and the loss function slightly decreases. To have a better understanding of the trained NKNs and GKNs, we look at the eigen-spectrum of their ``amplification matrices''. In particular, let the (discretized) $l-$th network layer be defined as $\mathbf{H}_{l}:=[h(x_1,l\Delta t),h(x_2,l\Delta t),\cdots,h(x_M,l\Delta t)]$, then the amplification from $\mathbf{H}_{l}$ to $\mathbf{H}_{l+1}$ can be written as as $\frac{\mathbf{H}_{l+1}-\mathbf{H}_{l}}{\Delta t} = \mathbf{A} \mathbf{H}_{l}+C\mathbf{1}$, where $\mathbf{A}$ is an $M\times M$ matrix, $C$ is a constant, and $\mathbf{1}$ is a size $M$ vector with all its elements equal to $1$. Note that since the kernel is layer-independent in both GKNs and NKNs, the amplification matrices $\mathbf{A}$ are also layer-independent. The analysis conducted in Theorem \ref{thm} tells us that if all eigenvalues of $\mathbf{A}$ are positive, the learnt operator is positive definite and the network is stable in the limit of deep layers. To test this fact, we randomly select a training sample pair $(f_j(x),u_j(x))$, extract the amplification matrices that connect subsequent trained layers, and compute their maximum and minimum eigen-values. These are reported in Table \ref{tab:eigen}; here, we observe that the NKNs' matrix is positive definite, which illustrates the theoretical results of Section \ref{sec:nkn}. In contrast, the GKNs' matrix exhibits negative eigenvalues, indicating that instabilities might occur.

\textit{Comparison between FNOs and NKNs:} Compared to GKNs and NKNs, FNO reaches a relatively low level of error on the training dataset ($O(10^{-3})$) when $L<32$. However, for $L\geq 32$, training FNOs becomes challenging due to the vanishing gradient phenomenon \cite{hochreiter1998vanishing}. From the right plot of Figure \ref{fig:poisson_loss} we can see that the test error of FNOs is $O(10^{-2})$; this values, being much larger than the training error, indicates that the network is overfitting the training data. This fact is possibly due to the fact the number of parameters increases with $L$. In fact, as reported in Table \ref{tab:1d_param}, for a $L-$th layer NN, FNO requires $L$ times more parameters than GKN and NKN. In contrast, NKNs trained with the shallow-to-deep initialization are robust and not subject to overfitting issues. Furthermore, FNOs proves to be more sensitive to the distribution of the training samples: without normalization, the test error increases by $10$ times. In contrast, regardless of normalization, NKNs reach the lowest test errors when $L>1$.

\subsubsection{Example 2: 2D Darcy's equation} 

\begin{table}[]
    \centering
    \begin{tabular}{|c|c|c|c|c|c|c|}
    \hline
    Model & $L=1$ &$L=2$&$L=4$&$L=8$&$L=16$&$L=32$ \\
    \hline
    GKN  & 473.20k & 473.20k & 473.20k & 473.20k & 473.20k & 473.20k \\
    NKN & 945.31k & 945.31k & 945.31k & 945.31k & 945.31k & 945.31k\\
    FNO &  171.42k & 338.37k & 672.26k & 1.34M & 2.68M & 5.35M\\
    \hline
    \end{tabular}
    \caption{Example 2: 2D Darcy's equation. Number of trainable parameters for each model.}
    \label{tab:2d_param}
\end{table}

We consider the two-dimensional heterogeneous PDE describing Darcy's flow and follow the same settings as in paper \cite{li2020neural} where GKNs are utilized. Here, the physical domain is $D=[0,1]^2$, the operator ${\rm L}_b$ is an elliptic operator with Neumann boundary conditions and permeability coefficient $b(\xb)$. We have:
\begin{align*}
-\nabla\cdot(b(\xb)\nabla u(\xb))=f(\xb),&\quad\xb\in D,\\
u(\xb)=0,&\quad\xb\in\partial D.
\end{align*}
We aim to learn the operator mapping from the parameter function $b(\xb)$ to the solution $u(\xb)$. As is standard in simulations of subsurface flow, the permeability $b(\xb)$ is modeled as a two-valued piecewise constant function with random geometry such that the two values have ratio 4. It is generated randomly for every sample and it is defined as $\psi_{\#}\mcN(0,(-\Delta+9I)^{-2})$, where $\psi$ takes the value 12 on the positive part of the real line and 3 on the negative. Different resolutions of data sets are down-sampled from a $241\times 241$ grid solution generated by using a second-order finite difference scheme. Training and validation are performed on the benchmark data set provided in \cite{li2020neural}; the corresponding data can be found at \url{https://github.com/zongyi-li/graph-pde}. We consider two training data sets: a ``coarse'' data set with grid size $\Delta x=1/15$ and hence $M=16\times 16$, and a ``fine'' data set with grid size $\Delta x=1/30$ and correspondingly $M=31\times 31$. With the purpose of testing generalization properties with respect to resolution, we consider three testing data sets: a ``coarse'' data set with grid size {$\Delta x=1/15$}, a ``fine'' data set with grid size {$\Delta=1/30$}, and a ``finer'' data set with grid size {$\Delta=1/60$}. 100 training samples and 40 test samples are employed. We again report learning results on both normalized (denoted as the ``with normalization'' cases) and original (denoted as the ``w/o normalization'' cases) training data sets.

For this example, we set the dimension $d$ of $\hb$ equal to 64. Following \cite{li2020neural}, we initialize $\hb(\xb,0)$ as 
\begin{equation}\label{eqn:init2D}
\hb(\xb,0) = P(\xb,b(\xb),b_{\epsilon}(\xb),\nabla b_{\epsilon}(\xb))+{\pb},
\end{equation}
where $P\in\real^{64\times 6}$, ${\pb}\in\real^{64}$, and $b_{\epsilon}(\xb)$ is a Gaussian smoothed version of the coefficients $b(\xb)$ obtained with a centered isotropic Gaussian distribution of variance $5$; $\nabla b_{\epsilon}(\xb))$ is its gradient. For an $L-$layer network, we apply \eqref{eq:NKN} iteratively, with $k(\xb,\yb,b(\xb),b(\yb)):\real^6\rightarrow \real^{4096}$ parameterized as a 3-layer feed forward network with widths $(6,512,1024,4096)$ and ReLU activation function. Note that the output of the network is then reshaped so to obtain a 64$\times$64 tensor. The domain of integration is restricted to the ball $B_r(\xb)$, with interaction radius $r = 0.10$, i.e., each node $\xb$ is only connected to nodes within distance $r$. The reaction network $R(\xb):\real^2\rightarrow \real^{4096}$ is parameterized as a 3-layer feed forward network with widths $(2,512,1024,4096)$ and ReLU activation. Also in this case, the output of the network is reshaped so to obtain a 64$\times$64 tensor. The network is trained with the shallow-to-deep training procedure. For each depth $L$, we initialize the network parameters from the $(L/2)-$layer NKN model, then train the network for {1000} epochs with a learning rate of $1e{-4}$, then decrease the learning rate with a ratio $0.8$ every 50 epochs.

\begin{figure}
    \centering
   \includegraphics[width = .7\textwidth]{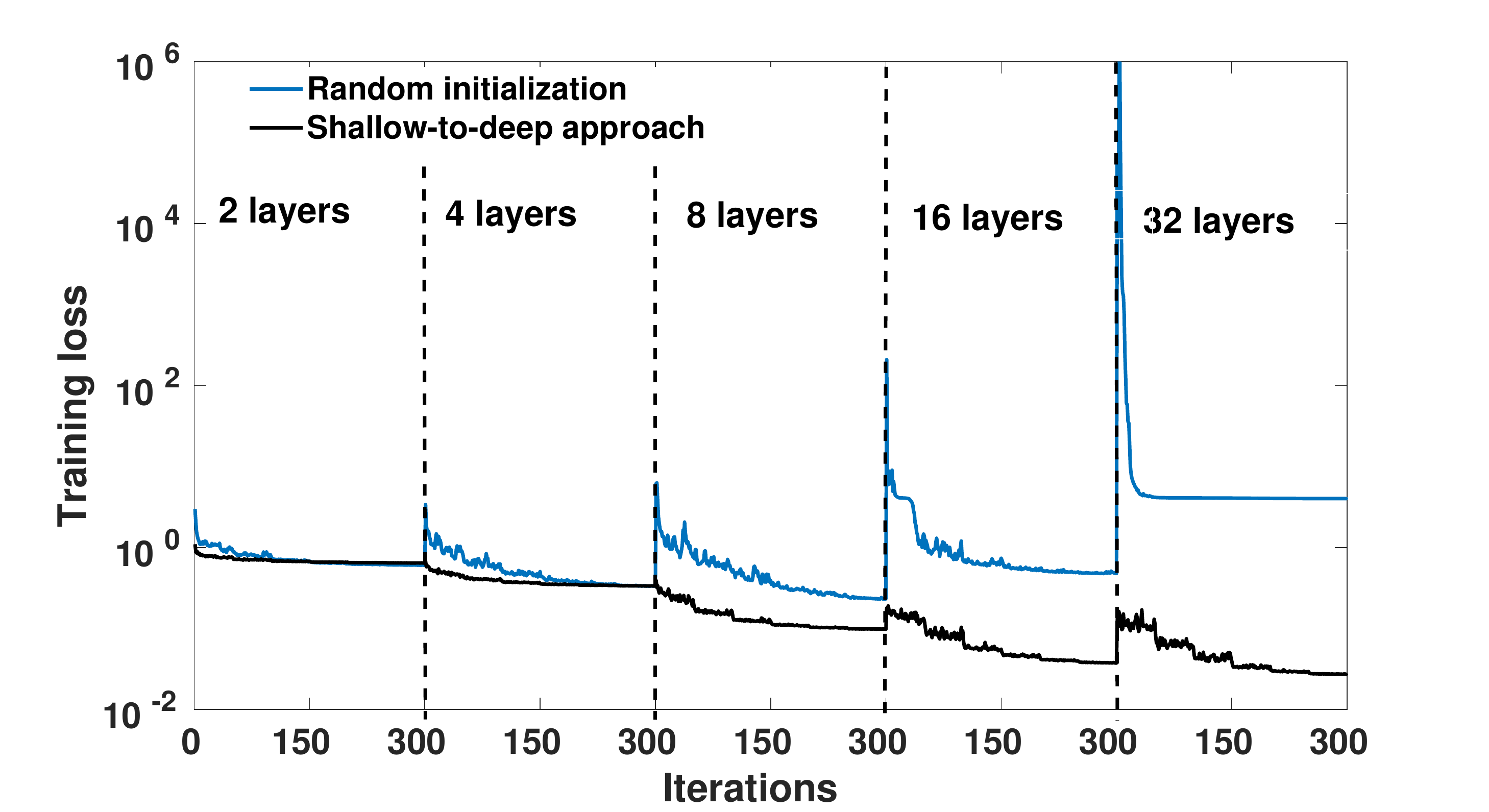}
    \caption{Example 2: 2D Darcy's equation. Training loss of the 2D Darcy problem using random initialization and shallow-to-deep approach, from 2 layers to 32 layers.}
    \label{fig:loss_idea3_2D}
\end{figure}

\begin{figure}
    \centering
    \includegraphics[width=0.48\textwidth]{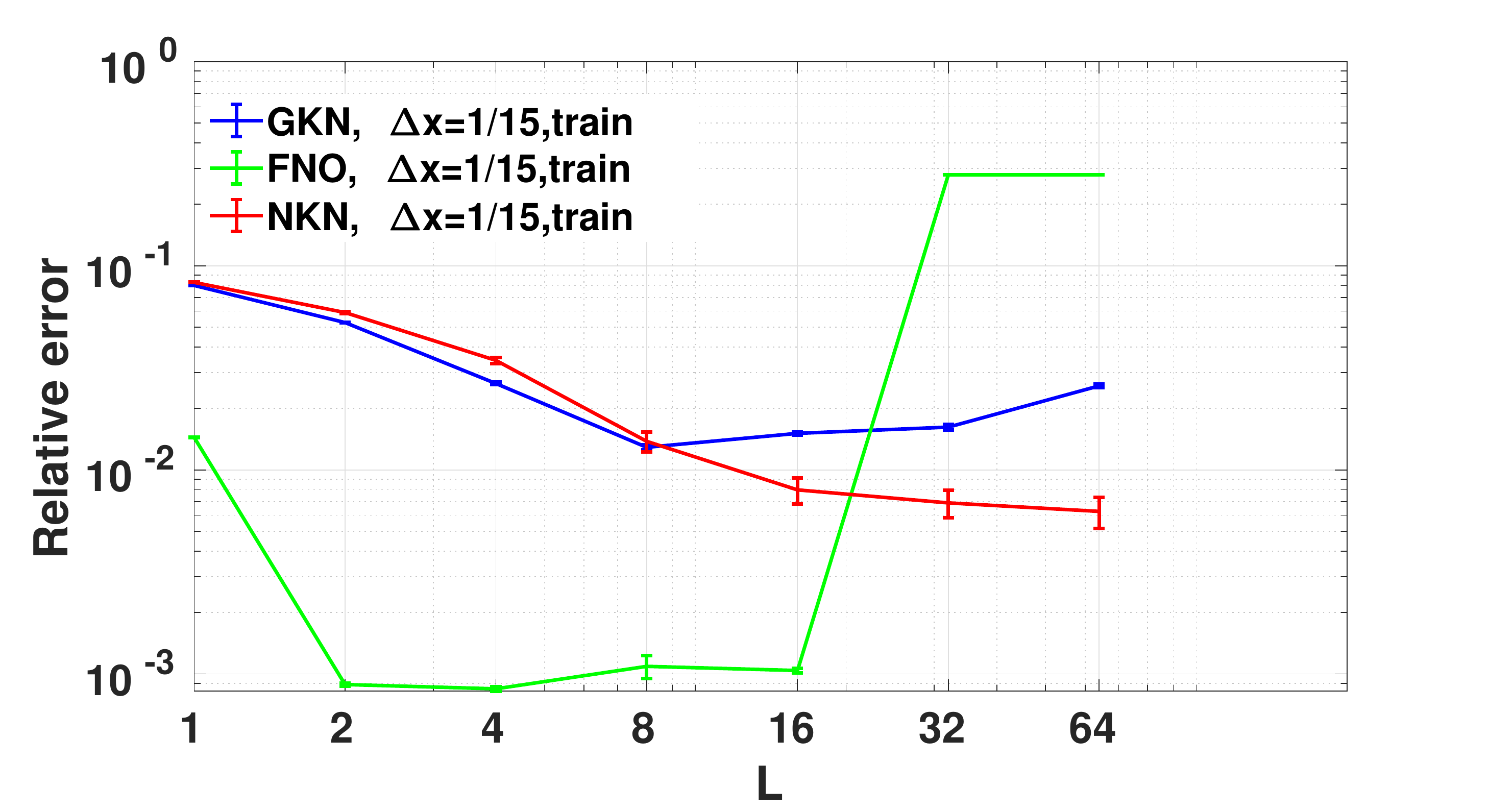}
    \includegraphics[width=0.48\textwidth]{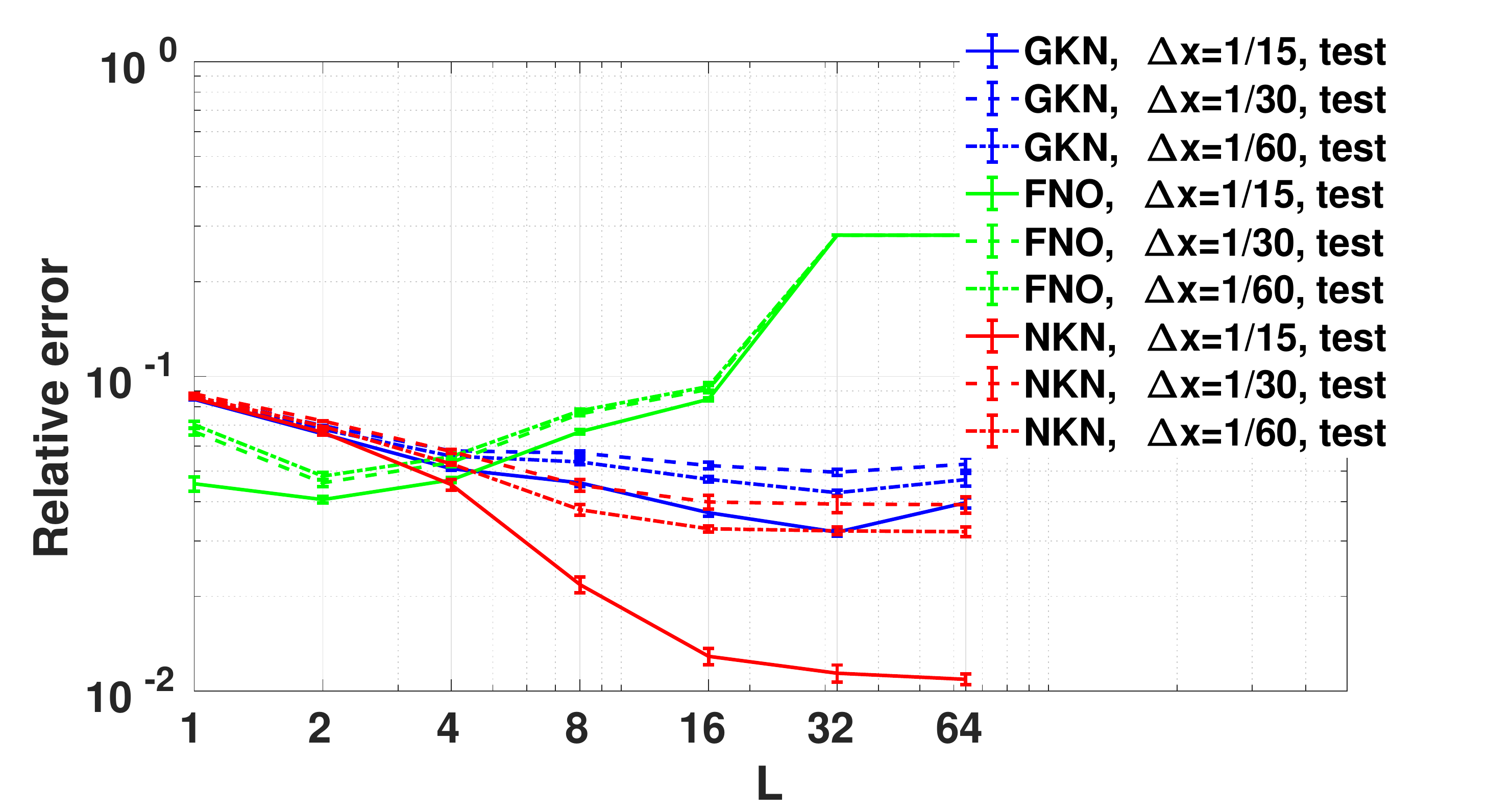}
    \includegraphics[width=0.48\textwidth]{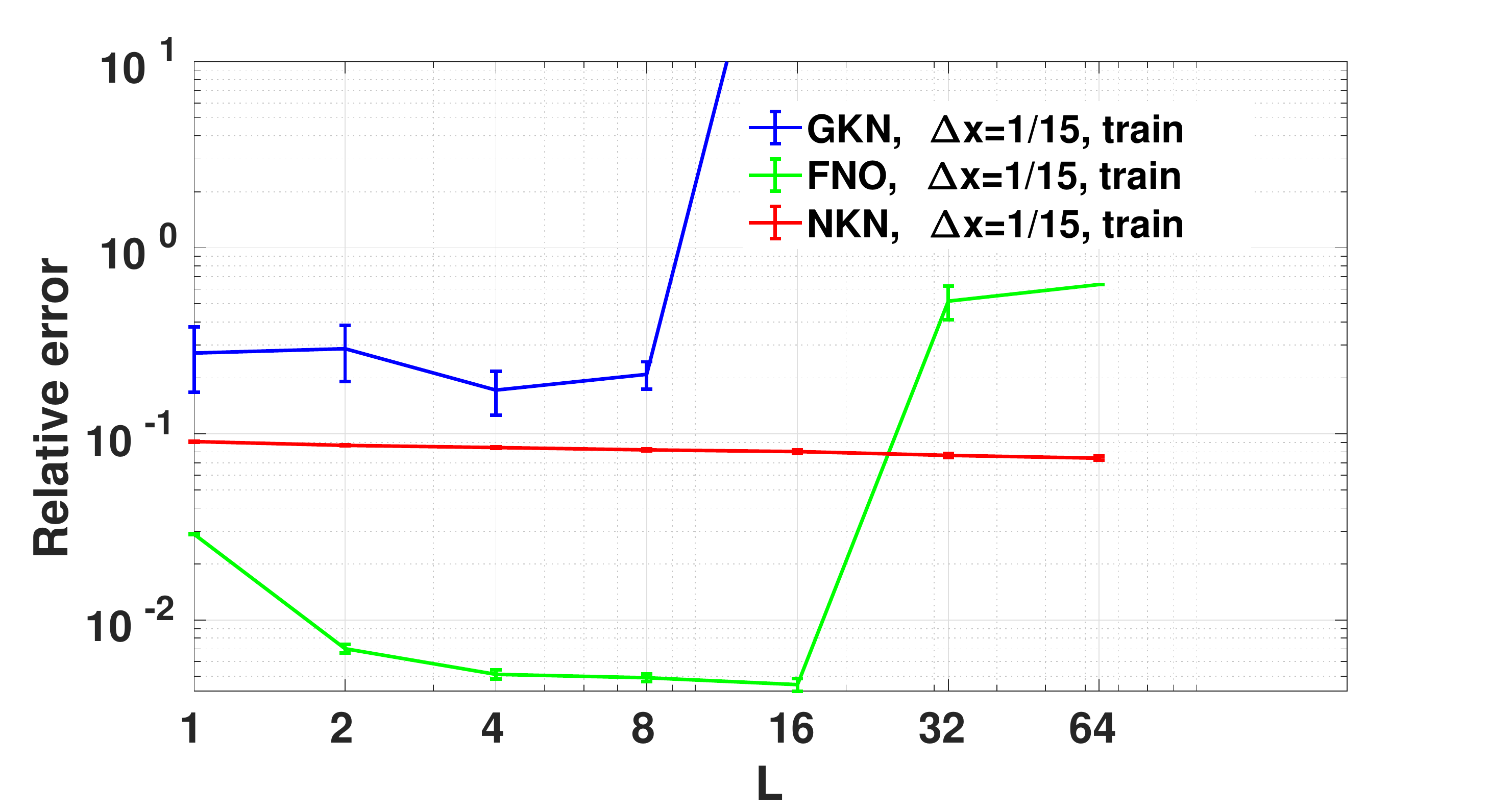}
    \includegraphics[width=0.48\textwidth]{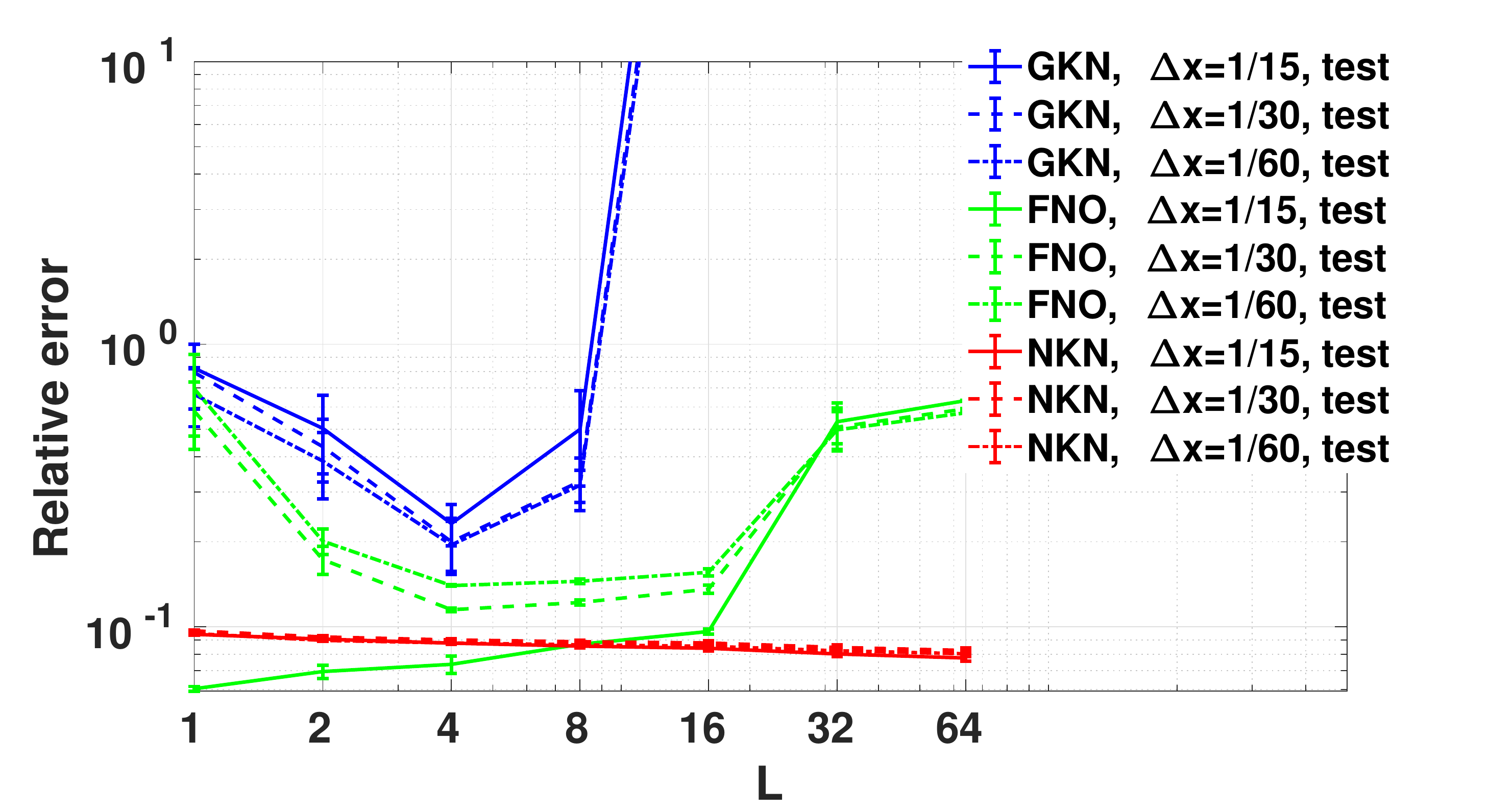}
    \caption{
    Example 2: 2D Darcy's equation. Comparison of relative mean squared errors from GKNs, FNOs, and NKNs when using the ``coarse'' training set ($\Delta x=1/15$). Error bars represent standard errors over 5 simulations. Top plots: training with the normalized dataset. Bottom plots: training with the original dataset. Left column: errors on the training dataset. Right column: errors on the test dataset with different resolutions.}
    \label{fig:loss_2ddarcy_16}
\end{figure}

\begin{figure}
    \centering
    \includegraphics[width=0.48\textwidth]{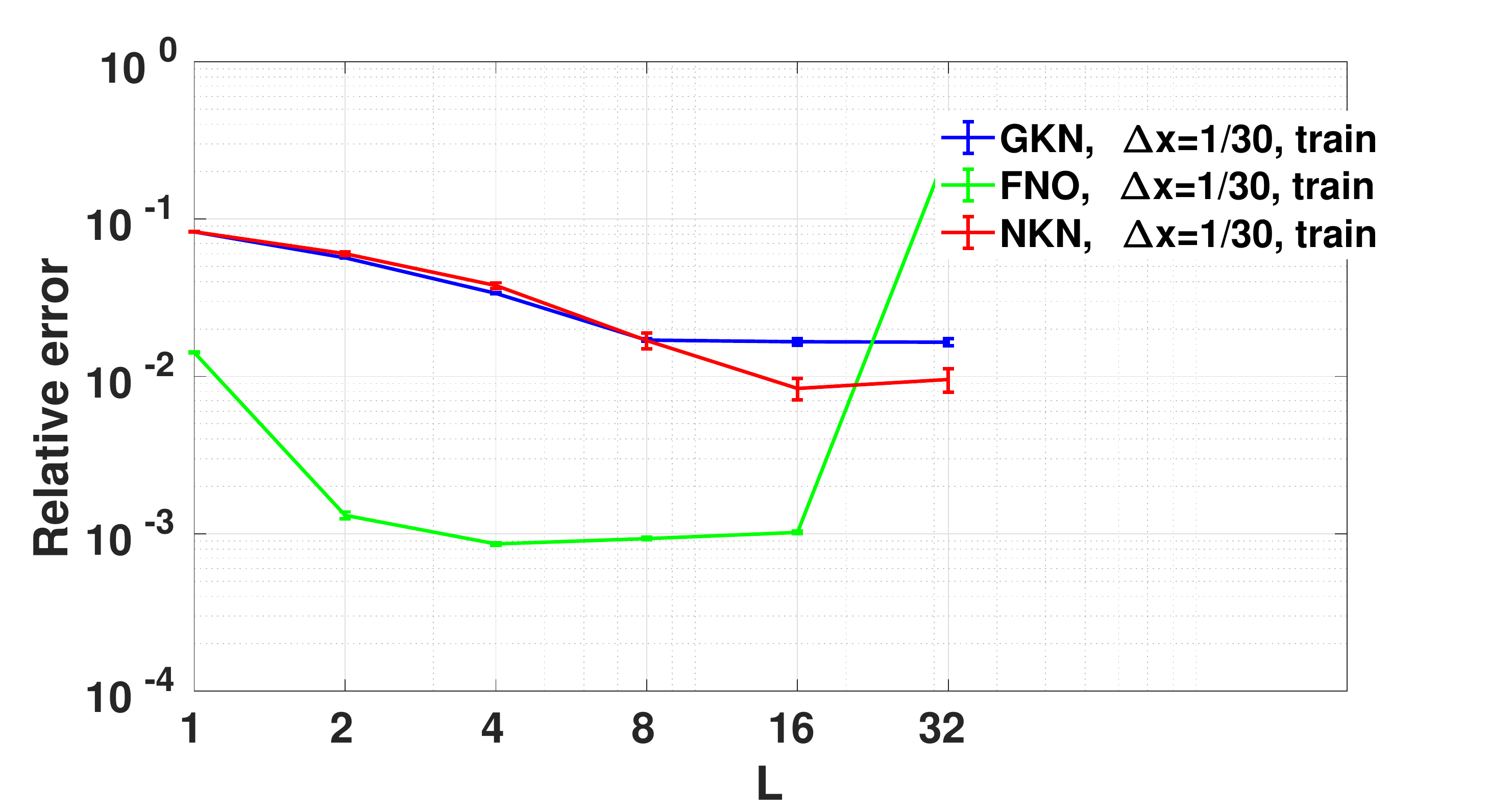}
    \includegraphics[width=0.48\textwidth]{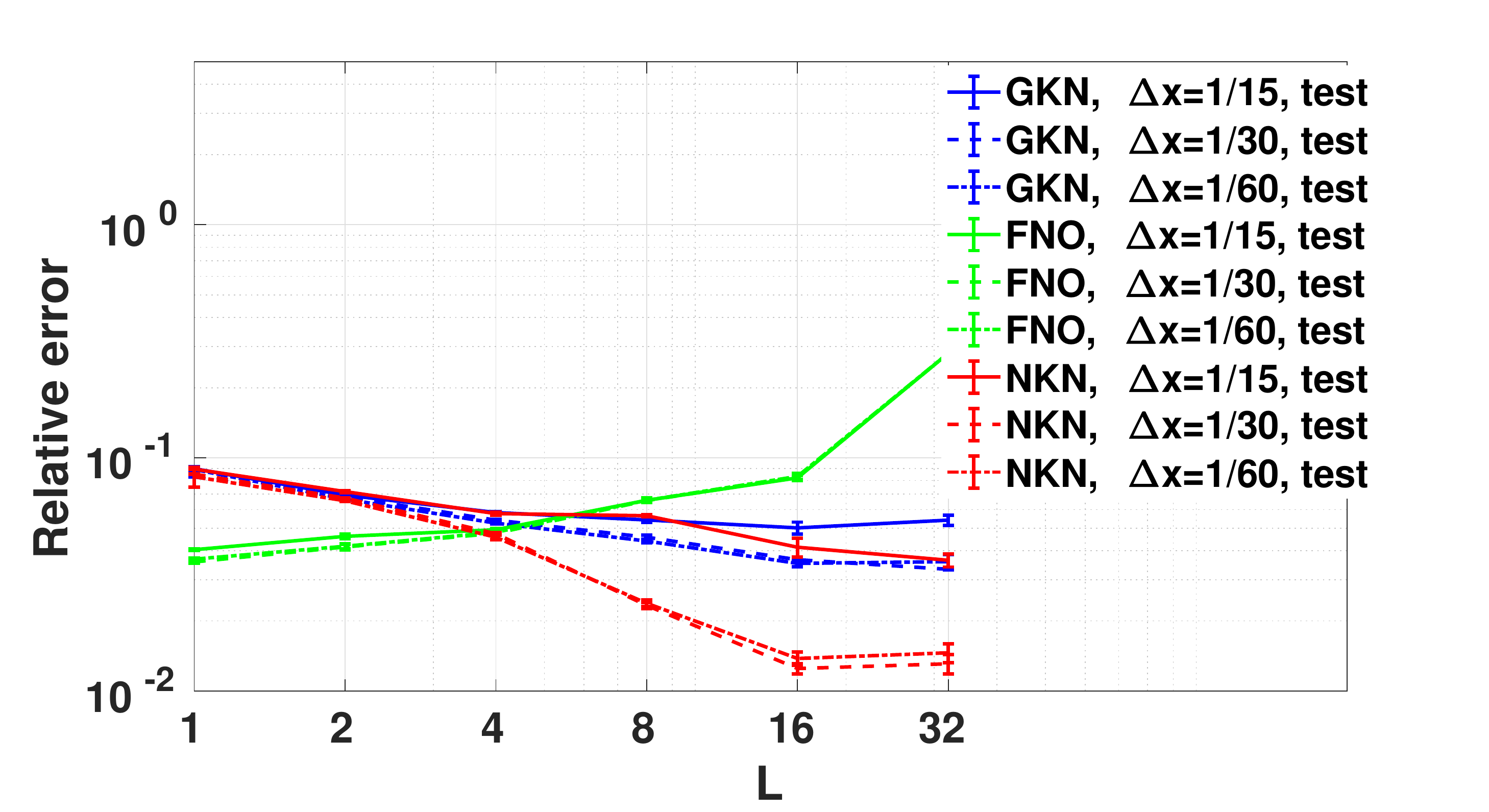}
    \includegraphics[width=0.48\textwidth]{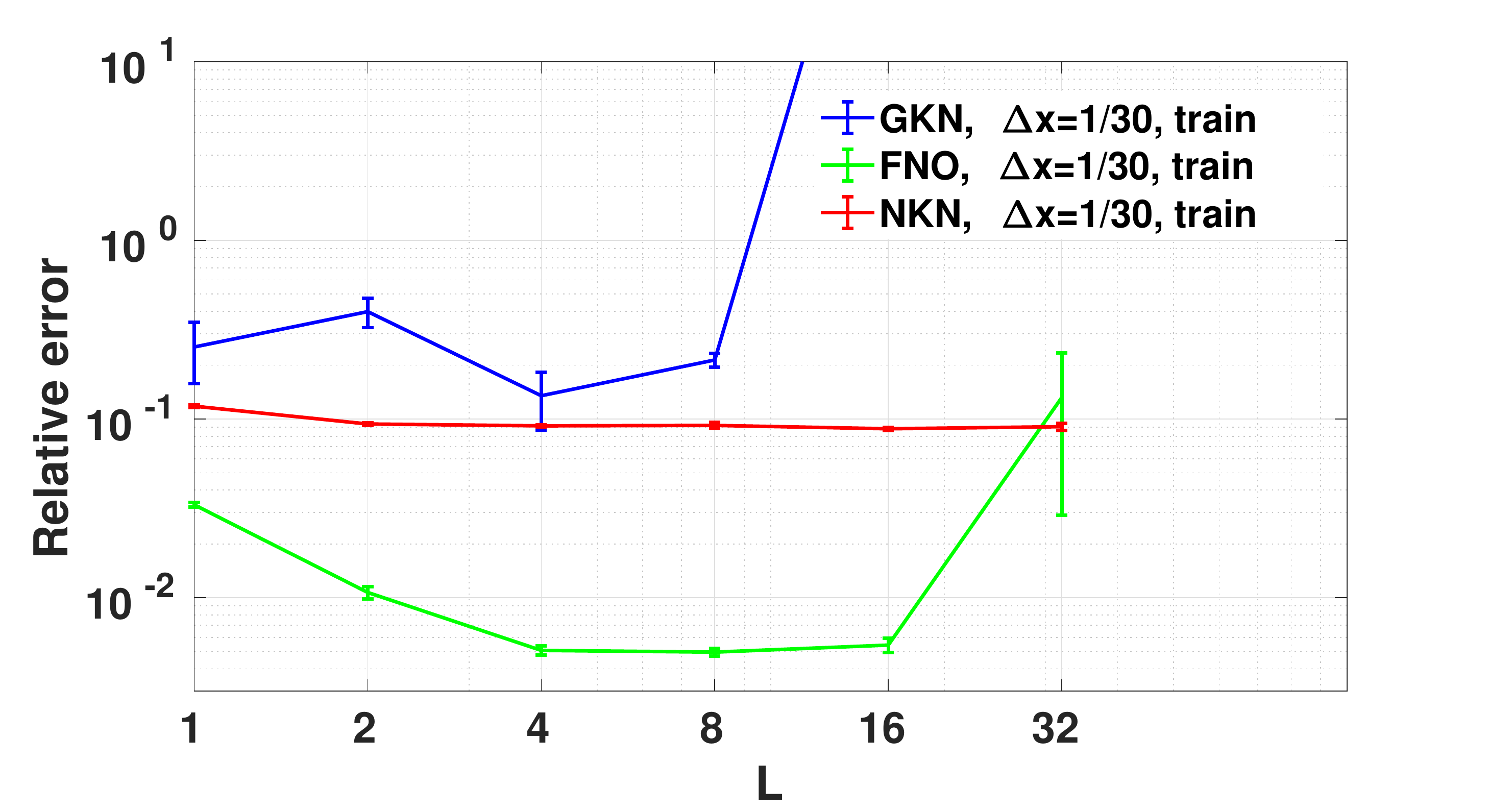}
    \includegraphics[width=0.48\textwidth]{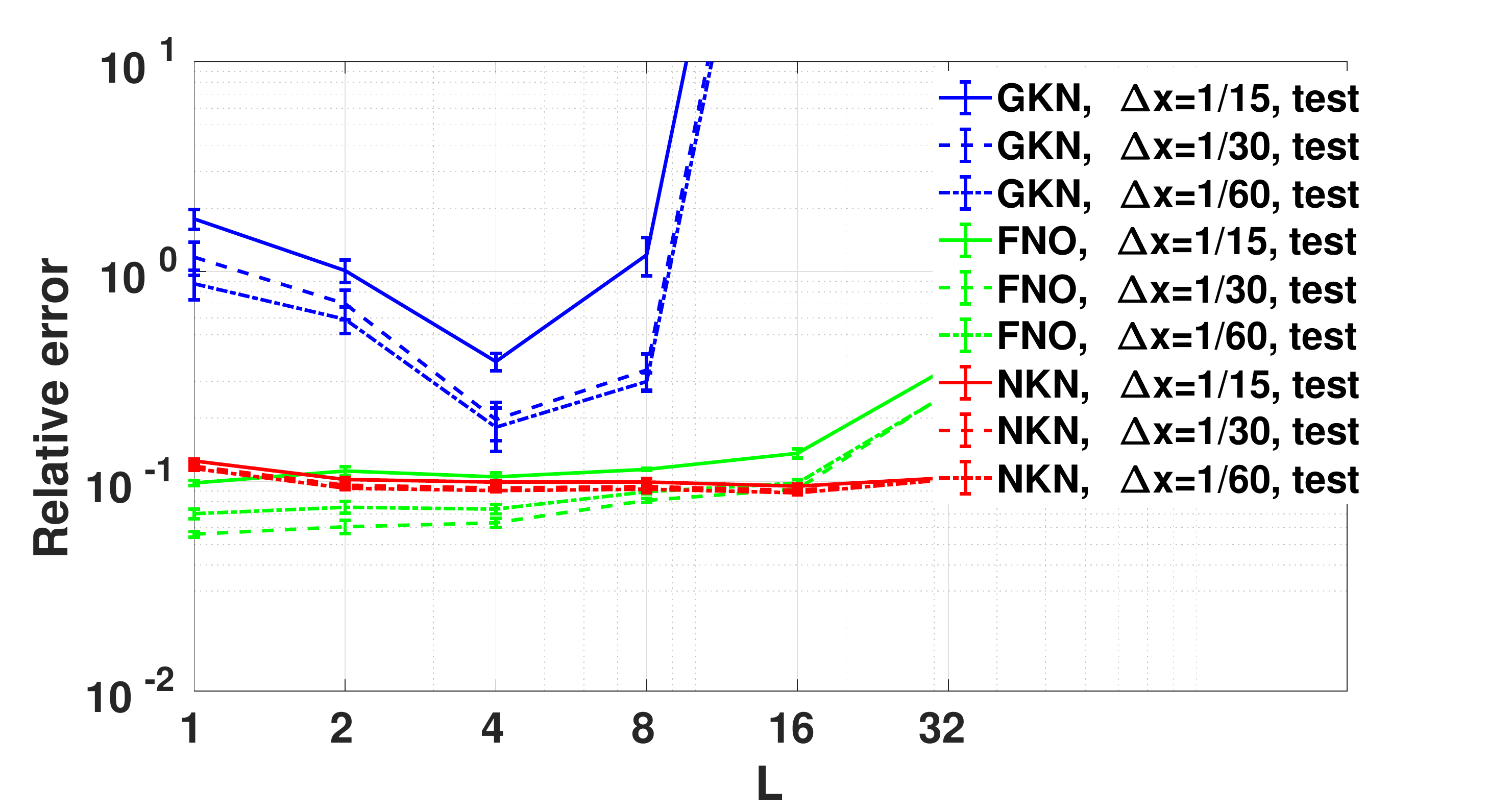}
    \caption{
    Example 2: 2D Darcy's equation. Comparison of relative mean squared errors from GKNs, FNOs, and NKNs when using the ``fine'' training set ($\Delta x=1/30$). Error bars represent standard errors over 5 simulations. Top plots: training with the normalized dataset. Bottom plots: training with the original dataset. Left column: errors on the training dataset. Right column: errors on the test dataset with different resolutions.}
    \label{fig:loss_2ddarcy_s31}
\end{figure}

\textit{Effect of the shallow-to-deep technique.} To illustrate the benefits of the shallow-to-deep initialization strategy, in Figure \ref{fig:loss_idea3_2D} we compare the convergence properties of the learning algorithm using random initialization and the shallow-to-deep initialization with $s=16$. Here, we successively double the number of layers from 2 to 32\footnote{For illustration we show the training loss with $300$ epochs for each depth $L$ in Figure \ref{fig:loss_idea3_2D}, although in the rest of this section we use $1000$ epochs to guarantee that each model has reached a plateau.}. The training losses are plotted with respect to the number of epochs. It can be seen that the initial guesses provided by the last network correspond to a lower value of the loss function. Therefore not only we have faster convergence, but we can also reach lower loss values. This is particularly important for deeper-layer networks, which are notoriously difficult to train and for which random initialization fails to provide accurate answers. This, we conclude that the shallow-to-deep technique provides a good initialization and an improved accuracy, which also helps avoiding the vanishing gradient issue in training.

\textit{Comparison between GKNs, FNOs and NKNs.} In Figures \ref{fig:loss_2ddarcy_16} and \ref{fig:loss_2ddarcy_s31}, we report the relative mean squared errors from the ``coarse'' and ``fine'' training data sets, respectively. Similarly to the Poisson's case, when increasing the number of layers, the relative training errors of GKNs and FNOs deteriorates for $L>8$, after initially decreasing. In contrast, the accuracy of NKNs monotonically improves\footnote{The only exception is for $\Delta x=1/30$ and $L=32$, where the training loss slightly increases from $L=16$, because we had to decrease the batch size in training, due to GPU memory constraints.} for increasing values of $L$. Also in this case, FNOs suffer from the overfitting: the test error increases as FNOs get deeper. Instead, when $L>4$, NKN consistently outperforms GKNs and FNOs in the testing experiments. Thus, while GKNs and FNOs remain reasonable choices when the network is at most 4 layers deep, NKNs achieve a better accuracy when the network is deeper than $4$. On the other hand, differently from Example 1, in this example normalizing the training data set helps improving the test error for all three architectures. In particular, for GKNs, the original training data set yields severe instabilities: the training loss blows up when $L>8$. For FNOs and NKNs, normalization also helps improving the test error. However, NKNs are still reliable when normalization is not performed. This fact becomes particularly important in online training, where normalization is not an option. In what follows, we always focus on the normalized case, unless otherwise stated.

\begin{figure}
    \centering
    \includegraphics[width=.8\textwidth]{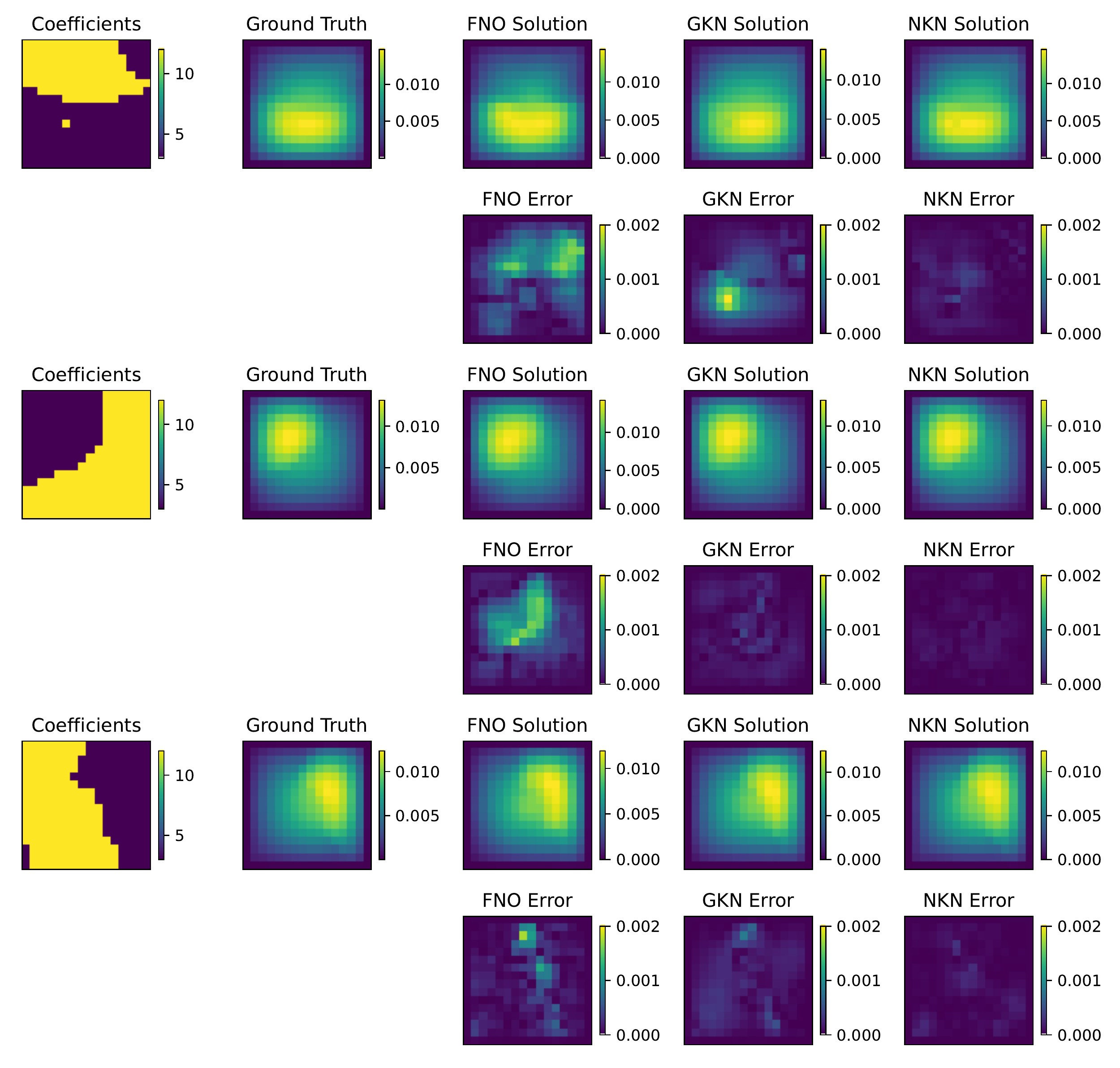}
    \caption{Example 2: 2D Darcy's equation. A visualization of 16-layer FNO, GKN, and NKN performance on three instances of permeability parameter $b(\xb)$, when using (normalized) ``coarse'' training dataset ($\Delta x=1/15$) and test on the dataset with the same resolution.}
    \label{fig:s16_test}
\end{figure}

To provide a qualitative comparison between GKNs, FNOs and NKNs, in Figure \ref{fig:s16_test} we show plots of solutions obtained with a 16-layer NKN, GKN, and FNO in correspondence of three instances of permeability parameter $b(\xb)$. For all cases the model is trained on the ``coarse'' data set and tested on the same resolution. Both the solutions and the errors are plotted. One can observe that all solutions obtained with NKN are visually consistent with the ground-truth solutions, while GKN loses accuracy near the material interfaces. FNO results are off in a even larger regions. These results provide further qualitative demonstration of the superiority of NKNs and confirm the conclusion inferred from the comparison in Figure \ref{fig:loss_2ddarcy_16}. For this case, the relative test errors for GKN, FNO and NKN are $3.69e-2\pm 9.28e-4$, $8.46e-2\pm1.03e-3$, $1.29e-2\pm7.61e-4$, respectively.

\begin{figure}
    \centering
    \includegraphics[width=.8\textwidth]{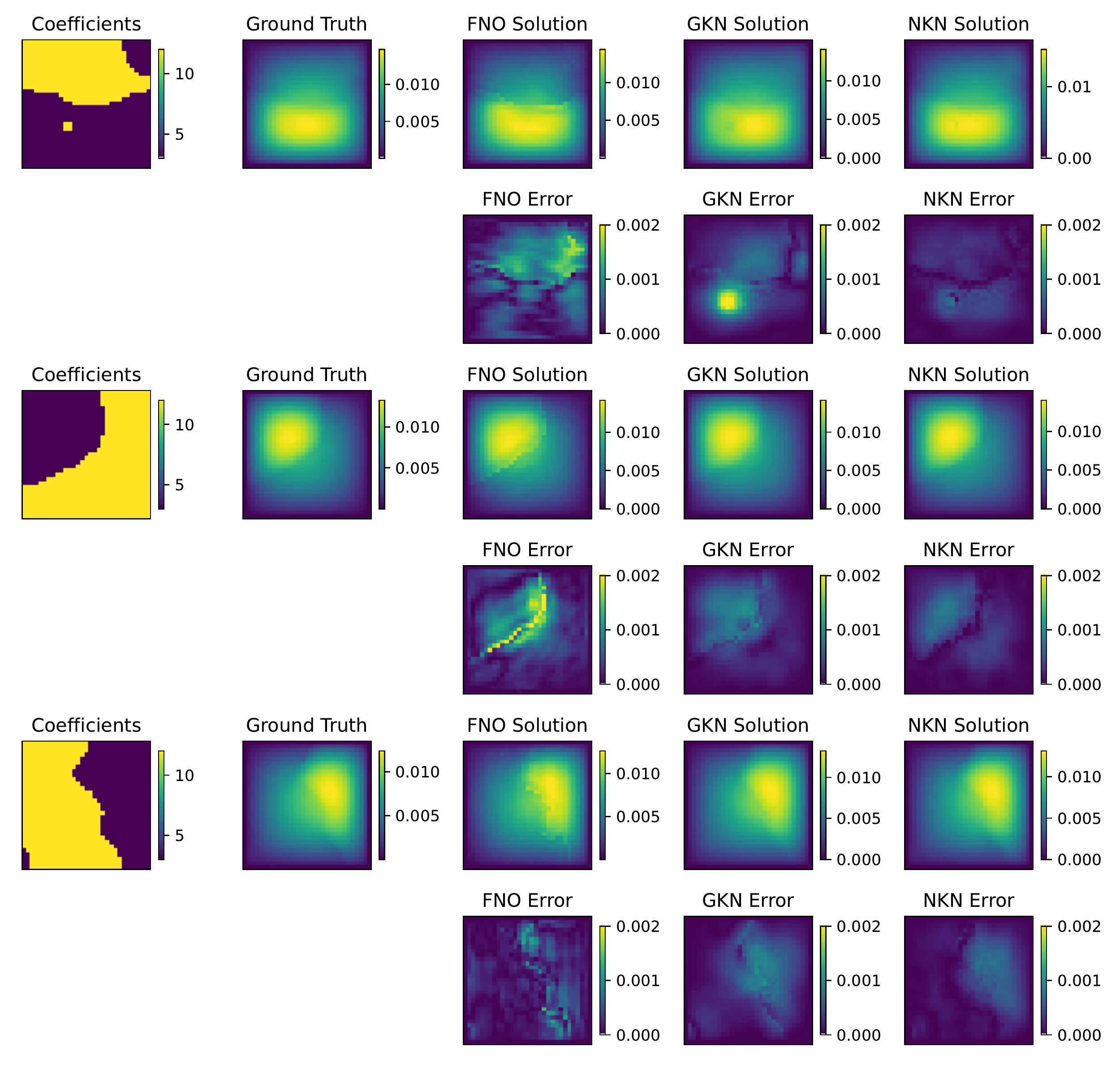}
    \caption{PDE learning task 2: 2D Darcy's equation. A visualization of 16-layer FNO, GKN, and NKN performance on three instances of permeability parameter $b(\xb)$, when using (normalized) ``coarse'' training dataset ($\Delta x=1/15$) and test on the ``fine'' dataset ($\Delta x=1/30$).}
    \label{fig:s31_test}
\end{figure}

\begin{figure}
    \centering
    \includegraphics[width=.8\textwidth]{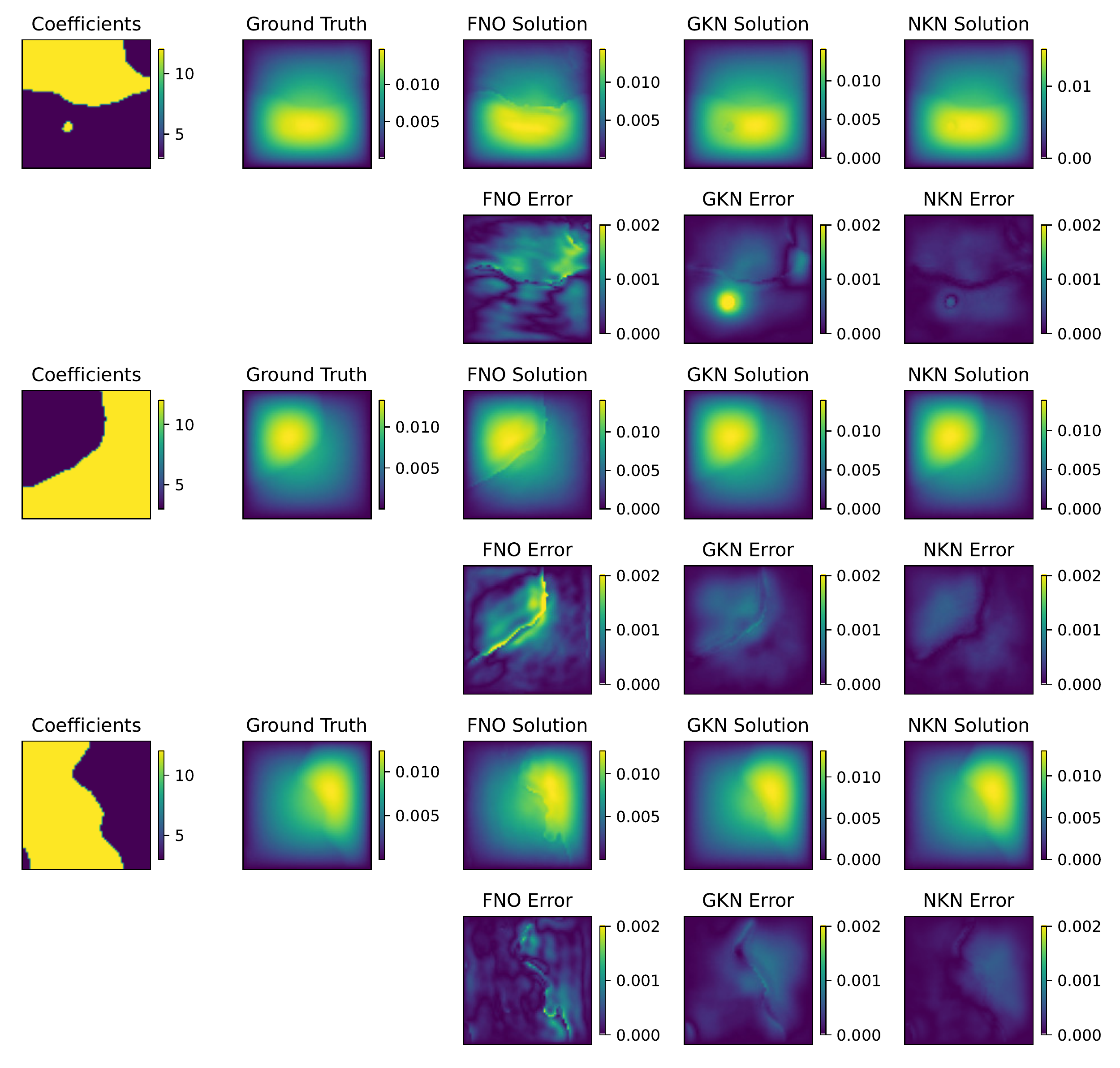}
    \caption{PDE learning task 2: 2D Darcy's equation. A visualization of 16-layer FNO, GKN, and NKN performance on three instances of permeability parameter $b(\xb)$, when using (normalized) ``coarse'' training dataset ($\Delta x=1/15$ and test on the ``finer'' dataset ($\Delta x=1/60$).}
    \label{fig:s61_test}
\end{figure}

\textit{Generalization to different resolutions.} To illustrate the generalization properties of GKNs, FNOs and NKNs to different grid resolutions, we train them with samples from a grid with resolution {$\Delta x=1/(s-1)$} and test them on samples from a grid with resolution {$\Delta x=1/(s'-1)$}. Test errors are provided in the right columns of Figures \ref{fig:loss_2ddarcy_16} and \ref{fig:loss_2ddarcy_s31}. 
We can observe that for each fixed training resolution $s$, the test errors at different resolutions remain on a similar scale for all three methods. We observe that when training on a grid of resolution {$\Delta x=1/30$}, the test error is smaller when the network is tested on resolution {$\Delta x=1/60$} than on {$\Delta x=1/15$}, indicating that testing on a fine grid provides better results. 
This is due to the fact that, for smaller $\Delta x$, the support of the kernel includes more grid points, leading to a better numerical integration. Instead, when utilizing the learnt network on a coarser resolution, the kernel is more likely to become less accurate, especially when the interaction radius $r$ is small. This observation was also reported in \cite{li2020neural}. A similar phenomenon is observed in image classification tasks, as further discussed in Section \ref{section:img}. 

To provide a visual comparison of the cross-resolution learning results, in Figures \ref{fig:s31_test} and \ref{fig:s61_test} we test the architectures trained with {$\Delta x=1/15$} on two data sets corresponding to {$\Delta x=1/30$} and {$\Delta x=1/60$}, and report the results for the same three instances of $\bb(\xb)$. It is again observed that NKNs outperform both baseline methods. 
We conclude this section stressing once again that the resolution-independence of these neural operators only guarantees that the generalization error is of the same order of the training error, i.e. when utilizing the operator to predict the solution associated to an input parameter on a finer (coarser) grid, the accuracy does not improve (worsen). For example, when utilizing NKNs trained with $\Delta x=1/15$ to predict inputs characterized by {$\Delta x=1/15$, $\Delta x=1/30$, and $\Delta x=1/60$}, we observe that the testing errors are of the same order, i.e. $1.29e-2\pm 7.41e-4$, $3.99e-2\pm2.02e-3$, and $3.28e-2\pm7.36e-4$, respectively.



\subsection{Image Classification Tasks}\label{section:img}

We illustrate the stability and resolution independence of NKNs using two supervised image classification problems. Specifically, we classify low-resolution images using networks trained on high-resolution images and vice-versa. Two benchmark image data sets are considered: the MNIST data set \cite{lecun1995learning} of handwritten digits available at \url{http://yann.lecun.com/exdb/mnist/}, and the CIFAR-10 data set \cite{krizhevsky2009learning} available at \url{https://www.cs.toronto.edu/~kriz/cifar.html}. This task corresponds to identifying the solution operator that maps the original image (represented by a discretized pixel valued function $\bb(\xb)$, where $\xb$ is the pixel location) to a vector-valued function $\ub(\xb)$ which represents the feature of this image. The class of the image will be obtained by applying a softmax classifier to $\ub(\xb)$. A resolution-independent map is such that it is equally accurate when classifying images $\bb$ with resolutions different from the training one.

We proceed as follows: given an image sample, we project it into the feature space by applying the transformation $\hb(\xb,0)=P(\xb,\bb(\xb))+{\bf p}$, where $\xb=(i,j)$, $i,j\in\mathbb{N}$, represents the pixel location and $\bb(\xb)$ is the initial pixel value at $\xb$. Then, we iteratively apply \eqref{eq:NKN}, 
$$
\hb(\xb,t+\Delta t)=\hb(\xb,t)+\Delta t\left(-R(\xb)\hb(\xb,t)+\int_{B_r(\xb)} k(\xb,\yb;\vb)(\hb(\yb,t)-\hb(\xb,t)) d\yb+\cb\right),
$$
and finally calculate the output feature function $\ub(\xb)=Q\hb(\xb,T)+{\bf q}$ and the predicted class of the given image sample as  $\text{softmax}(\ub(\xb))$. Note that, in the integral above, to accelerate the training we restrict the domain of integration to a neighborhood. In other words, each node $\xb$ is only connected to nodes within a distance $r$, i.e. to nodes in the neighborhood $B_r(\xb):=\{\yb:|\yb-\xb|<r\}$. In all image classification tasks, we set the dimension $d$ of $\hb$ equal to 16, and the inner kernel network $k$ to be a 3-layer feed forward network with widths $(4,32,32,256)$ and ReLU activation function.  $R$ is also a 3-layer feed forward network with widths $(2,32,32,256)$ and ReLU activation. Both $k$ and $R$ are then reshaped into tensors of size 16$\times$16. Note that in image classification tasks, the network update above is often added to standard ResNet architectures, rather than utilized as a stand-alone network. This technique was also used in \cite{tao2018nonlocal} to enhance the accuracy of ResNets. Thus, in this case, $\bb$ may also represent the output of the previous ResNet layer.

In this section we compare NKNs to three baseline methods: CNNs \cite{albawi2017understanding}, multiscale CNNs \cite{haber2018learning}, and NNNs \cite{tao2018nonlocal}. For CNNs, we consider the standard convolution kernels of dimension $3 \times 3 \times 16$ and ReLU activation functions. After $L$ layers, we connect the output with another dense layer of output dimension 128 and a ReLU activation function, and finally connected to a softmax classifier. In the cross-resolution tests, we do not change any trained parameter nor the CNN kernels. For the multiscale CNN, we follow \cite{haber2018learning} and employ the same CNN structure, with a tanh activation function instead of the ReLU activation function for the CNN layers. In the cross-resolution tests, two transformation matrices are employed: a prolongation matrix $\mathbf{S}$ that maps coarse images into higher resolutions and a restriction matrix $\mathbf{U}$ that performs the opposite mapping. $\mathbf{S}$ is given by a bilinear interpolation and constant padding.
$\mathbf{U}$ maps a fine image into a coarse image in such a way that $\mathbf{U}\mathbf{S} = \mathbf{I}$, the identity operator. Note that the CNN layer on a fine image can be viewed as a linear operator and rewritten as a sparse matrix $K_h$. Therefore, when using CNNs trained with fine images on coarse images, the convolution operator is adjusted to the coarse scale as $K_H = \mathbf{U}K_h\mathbf{S}$. When applying CNNs trained with coarse images on fine images, CNN layers are similarly adjusted as $K_h = \mathbf{S}K_H\mathbf{U}$. For NNNs we follow the conventions in \cite{tao2018nonlocal}: the NNN's input layer is followed by a dense layer with 16 output dimensions. The iterative formulation \eqref{eq:nnn} is then employed, followed by another dense layer of output dimension 128 and a ReLU activation function, and finally connected to a softmax classifier. We use the Adam optimizer to train all these baseline models until a plateau is reached (often within 200 epochs).

\subsubsection{Example 1: MNIST}

\begin{table}[]
    \centering
    \begin{tabular}{|c|c|c|c|c|}
    \hline
    \multirow{2}{*}{Model} & \multicolumn{2}{c|}{Trained on fine images}
    & \multicolumn{2}{c|}{Trained on coarse images} \\
    \cline{2-5}
    &  Validation(fine) & Validation(coarse) & 
    Validation(fine) & Validation(coarse) \\
    \hline
    CNN, $L=1$ & 2.55\% & 27.88\%   & 36.25\% & 3.23\% \\
    CNN, $L=2$ & 2.08\% & 28.75\%  & 48.35\%  & 2.09\%\\
    CNN, $L=4$ & {\bf1.68\%} & 90.42\%  & 37.14\%  & {1.84\%} \\
    \hline
    Multiscale CNN$^*$ (\cite{haber2018learning}) &1.82\%&5.08\%&9.98\%&{\bf1.72\%}\\
    Multiscale CNN, $L=1$ & 3.50\% & 49.21\% & 14.46\% & 4.22\% \\
    Multiscale CNN, $L=2$ & 2.46\% & 57.74\%  & 78.45\%  & 2.54\% \\
    Multiscale CNN, $L=4$ & 2.01\% & 56.56\% & 91.08\% &  1.84\% \\
    \hline
    NNN, $L=1$ & 4.31\% & 10.66\% & 9.51\% & 5.05\% \\
    NNN, $L=2$ & 4.48\%  &  9.58\% & 8.06\% & 4.63\% \\
    NNN, $L=4$ & 4.15\% & 11.27\% & 10.51\% & 4.72\% \\    
   \hline 
   NKN, $r=2$, $L=1$ & 3.37\% & {\bf 4.37\%} & 4.55\% & 4.53\% \\
   NKN, $r=2$, $L=2$ & 3.26\% & 4.98\% & 9.15\% & 4.35\% \\
   NKN, $r=2$, $L=4$ & 3.26\% & 4.51\% & 10.92\% & 4.29\% \\
     NKN,  $r=3$, $L=1$  & 3.40\% & 4.96\% & {\bf3.76\%} & 3.75\% \\
     NKN, $r=3$, $L=2$  & 3.20\% & { 4.85\%} &   4.02\% &  3.52\%\\
    NKN,  $r=3$, $L=4$ & 3.28\% & 5.87\% & 5.95\%  & 3.40\%  \\
    NKN,  $r=4$, $L=1$  & 3.28\% & 5.83\% & 5.37\%  &  3.94\%\\
     NKN, $r=4$, $L=2$  & 3.26\%  & 6.12\% &  4.90\% & 3.63\% \\
    NKN,  $r=4$, $L=4$ & 3.23\% & 5.48\% &   4.88\% & 3.58\%\\
      \hline 
    \end{tabular}
    \caption{Image classification task 1: MNIST. Image classification errors on test dataset (lower is better). Bold numbers highlight the best case. ``Multiscale CNN$^*$ \cite{haber2018learning}'' reports the values from \cite{haber2018learning}. ``$r=*$'' and ''$L=*$'' indicates the interacting radius in NKN and the number of CNN/NNN/NKN layers employed in the model, respectively.}
    \label{tab:mnist_more}
\end{table}

We first consider the MNIST data set which has a training set of 60,000 labeled images. These samples consist of $28\times 28$ black and white images and they will be employed as the fine-scale images. We randomly divide the data set into a training set consisting of 50,000 images, and a test set consisting of 10,000 images. In the cross-resolution classification task, images of two levels of resolutions are considered. We denote the original MNIST images as the ``fine images'', and generate ``coarse images'' by downsampling each image to a $14\times 14$ resolution using bilinear interpolation. We train two networks using the coarse and fine training data sets and then use the trained networks to classify both the fine and coarse validation data sets.

Results are reported in Table \ref{tab:mnist_more} for both the baseline architectures and NKNs. We point out that for Multiscale CNNs we show both the values reported in \cite{haber2018learning}, denoted by ``Multiscale CNN$^*$ \cite{haber2018learning}'' and the results from our implementation. Our Multiscale CNN results mostly differ from the ones in \cite{haber2018learning} in the cross-resolution test errors; this is due to the fact that non-standard loss functions (regression loss), different optimization methods (Block-Coordinate-Descent method), and additional regularization terms (derivative-based regularization term) are employed in \cite{haber2018learning}. Instead, in our setting, for a fair comparison with other methods, we employ the cross entropy loss and the Adam optimizer. The latter choices are standard in image classification tasks.

From Table \ref{tab:mnist_more} we can see that, while CNNs perform best when training and testing resolutions are the same, NKNs outperform other architectures when tested on a resolution different from the training one. In fact, when $r>2$, NKNs' testing errors at different resolutions are of the same order of the ones at the same resolution. This fact illustrates the resolution-independence property of NKNs. When the interaction radius $r$ is as small as $2$, NKNs are less accurate on cross-resolution tasks, although the overall test error is still of the same order as the training one and greatly outperforms the two baseline CNNs. This is due to the fact that when the interaction radius $r$ is too small, the support of the kernel contains only a small number of grid points, inducing a less accurate numerical integration. When comparing the NKN with $r=3$ and the NKN with $r=4$, we do not observe a significant improvement in accuracy as $r$ increases. This is possibly due to the fact that MNIST's data-label relation is relatively simple, so that $r=3$ is sufficient.

\subsubsection{Example 2: CIFAR}

We utilize the CIFAR-10 data set to illustrate the performance of NKNs in cross-resolution testing. CIFAR-10 consists of 50,000 training images and 10,000 test images of size $32\times 32$, belonging to ten classes. In this test, we consider three validation data sets containing images of three different resolution levels, following the same approach as in \cite{haber2018learning}. The ``original'' resolution data set consists of the original $32\times 32$ images, the ``fine'' resolution data set consists of $64\times 64$ images generated by bilinear interpolation, and the ``coarse'' resolution data set consists of $16\times 16$ images also generated by bilinear interpolation.

Differently from the approach used for MNIST in the previous section, and following the strategy described in \cite{tao2018nonlocal}, we incorporate the NKN network update (or nonlocal block) into a 20-layer pre-activation ResNet (PreResNet-20) \cite{he2016identity}. We compare NKNs with two baseline architectures: the standard PreResNet-20 with CNN blocks (denoted as ``baseline''), and NNNs where the nonlocal blocks (of depth $L=2,3,4,5$) are incorporated into the standard PreResNet-20 after the second residual block. Also for NKNs, we insert network updates into PreResNet-20 following the same procedure used for NNNs. To improve the descriptive power of NKNs, we employ different kernels $k$ at each layer, i.e., the kernel $k(\xb,\yb,t)$ and $R(\xb,t)$ are time-dependent functions. Therefore, the overall nonlocal network can be written as
$\hb(l+1):=\hb(l)+\mcF(\hb(l);W(l)),
$
where $W(l)$ is the parameter set, $l=0,1,\cdots,L_{total}$ with $L_{total}$ being the total number of network blocks. When the $l-$th block is nonlocal, we employ the architecture in \eqref{eq:NKN} and set $t=l\Delta t$ with 
$$\mcF(\hb(t+\Delta t)):=\Delta t\left(-R(\xb,t;\wb)\hb(\xb,t)+\int_D k(\xb,\yb,t;\vb)(\hb(\yb,t)-\hb(\xb,t)) d\yb+\cb(t)\right),$$
otherwise, the block is a traditional residual block of the pre-activation ResNet:
$\mcF(\hb(l)):=W_2^l g (W_1^l g(\hb(l))),$
where $g=\text{ReLU}\circ \text{BN}$ denotes the composition of ReLU and batch normalization (BN). The dimension of $\hb$ is set to $d=16$. For each NKN layer, the kernel network $k(\cdot,\cdot,t): \mathbb{R}^4 \rightarrow \mathbb{R}^{256}$ is parametrized as a 3-layer feed forward network with dimensions $(4, 32, 32, 256)$ and ReLU activation, and the reaction network $R(\cdot,t):\mathbb{R}^2 \rightarrow \mathbb{R}^{256}$ is parametrized as a 3-layer feed forward network with widths $(2, 32, 32, 256)$ and ReLU activation. Both are then reshaped into a 16$\times$16 tensor. As done for the MNIST data set, different radii $r = \{2,3,4\}$ are utilized. All models are implemented based on a 20-layer pre-activation ResNet (PreResNet) package in Keras provided in \cite{He2016Resnet} with default structure. Following the settings reported in \cite{tao2018nonlocal}, we set Adam's initial learning rate to $1e{-3}$, and train for 200 epochs.

Classification results are reported in Table \ref{tab:Image_CIFAR_more}. Here, for NNNs tested on the original resolution data set, we report both the results obtained with our implementation and the ones reported in \cite{tao2018nonlocal}. We observe that the performance of these two implementations is slightly different; this is possibly due to the differences in the Tensorflow version or in the available hardware. When testing on a data set with the original resolution, we can see that NKNs with $r=4$ and $4$ blocks outperform both the baseline and the best NNN. 

As for the cross-resolution classification tests, we train the networks using the $32\times 32$ images and then test their generalization properties on finer ($64\times 64$) and coarser ($16\times 16$) images. 
Due to the poor performance of CNNs in cross-resolution tasks (since they are formulated at the discrete level and hence not resolution-independent), when testing NNNs and NKNs on different-resolution images, we follow an approach similar to what we described for multiscale CNNs. Precisely, testing on finer images, the convolution operator $K_h$ is approximated by the trained convolution operator $K_H$ as $K_h = \mathbf{S}K_H\mathbf{U}$. If multiple CNNs are stacked together, we have 
$K_{h_n}K_{h_{n-1}}\cdots K_{h_1} = \mathbf{S}K_{H_n}K_{H_{n-1}}\cdots K_{H_1}\mathbf{U}$, since $\mathbf{S}\mathbf{U}= \mathbf{I}$. This is equivalent to multiplying by a restriction matrix $\mathbf{U}$ after the input layer, a prolongation matrix $\mathbf{S}$ before the NNN/NKN layer, and a restriction matrix $\mathbf{U}$ after the NNN/NKN layer. A similar procedure can be utilized when testing on coarser images; however, we expect results to be less accurate as $\mathbf{U} \mathbf{S} \neq \mathbf{I}$. We can see that among all architectures, NKNs are again the most accurate classifiers. Differently from what we observed for MNIST, here, NKNs are more accurate when a larger radius $r$ and a deeper network is employed. This is possibly due to the fact that the CIFAR-10 data set has a more complex data-label relation and therefore requires deeper architectures. Another interesting finding is that for all architectures it is easier to generalize to fine-scale images than to coarse-scale images. This is because when generalizing to a smaller grid, part of the support of the kernel is lost which causes the kernel to be inaccurate.

\begin{table}[]
    \centering
{    \begin{tabular}{|c|c|c|c|}
    \hline
    Model& Original/Reported in \cite{tao2018nonlocal} &Fine & Coarse  \\
    \hline
    Baseline&8.69\%/8.19\% & 50\% & 37.15\% \\
    \hline
    NNN, block $L=2$ & 8.04\%/7.74\% & 11.27\% & 38.15\% \\
    NNN, block $L=3$ & 8.09\%/7.62\% & 8.80\% & 28.18\% \\
    NNN, block $L=4$ & 8.10\%/7.37\% & 9.56\%  & 32.03\%  \\
    NNN, block $L=5$ (best)& 8.03\%/7.29\%& 11.86\% & 48.69\% \\
    \hline
    NKN, $r=2$, block $L=2$ & 7.94\%   & 8.10\%  & 46.86\% \\
    NKN, $r=2$, block $L=3$ & 7.60\% & 7.71\%  & 40.34\%\\
    NKN, $r=2$, block $L=4$ & 7.52\%  & 7.61\% & 40.28\% \\
    NKN, $r=3$, block $L=2$ & 7.60\% & 7.77\% & 24.81\% \\
    NKN, $r=3$, block $L=3$ & 7.67\%  & 7.78\%& 25.96\%  \\
    NKN, $r=3$, block $L=4$ & 7.94\% & 8.11\% & 26.78\% \\
    NKN, $r=4$, block $L=2$&7.70\% &  7.41\% & 31.80\% \\
    NKN, $r=4$, block $L=3$&7.23\%& 7.30\% & {\bf 23.16\%} \\    
    NKN, $r=4$, block $L=4$ & {\bf 7.08\%} &  {\bf 7.23\%} &  24.30\% \\
    \hline
    \end{tabular}}
    \caption{Image classification task 2: CIFAR-10. Image classification task errors. Bold numbers highlight the best case. For the baseline (PreResNet-20) and NNN cases, we report both the results from our implementation using the same hyperparameters and the ones reported in \cite{tao2018nonlocal}. For NNN and NKN cases, ``block $L=*$'' indicates the number of NNN/NKN layers employed in the inserted nonlocal block.}
    \label{tab:Image_CIFAR_more}
\end{table}

\section{Conclusion}\label{sec:conclusion}

We proposed a new integral neural operator, inspired by graph kernel networks, that has rigorous mathematical foundations provided by the nonlocal theory. This network, referred to as nonlocal kernel network (NKN), is stable in the deep network limit by construction. Similarly to neural ODEs, NKNs can be reinterpreted as time dependent equations. Furthermore, both layers and nodes are treated continuously. This fact, enables resolution independence and the use of efficient initialization techniques that exploit the continuous-in-time nature of NKNs. Our results show that, in both learning governing equations and image classification tasks, NKNs outperform baseline methods in stability and generalizability to different resolutions.

Similarly to GKNs, since NKNs' building blocks are integral operators characterized by space dependent kernels with minimal assumptions, they come at the price of higher computational cost compared to other networks whose kernels have a convolutional structure such as the standard CNN and FNO. 
However, since training cost can be seen as an offline cost, once the network is trained, prediction is a fast operation. Therefore, the excellent generalization properties of NKNs make them a valuable and robust tool for offline learning tasks and, due to the fact that they are insensitive to normalization, also for online learning tasks. Finally, NKNs represent one of the first examples of universal learning tools, being able to succeed in learning tasks of substantially different nature.


\section*{Acknowledgements}

 The authors would like to thank Dr. Yunzhe Tao and Dr. Zongyi Li for sharing their codes and for the helpful discussions. The authors also want to acknowledge Dr. Lars Ruthotto for providing implementation details regarding Multiscale CNN.
 
 H. You and Y. Yu would like to acknowledge support by the National Science Foundation under award DMS 1753031. Portions of this research were conducted on Lehigh University's Research Computing infrastructure partially supported by NSF Award 2019035. 
 
S. Silling and M. D'Elia would like to acknowledge the support of the Sandia National Laboratories (SNL) Laboratory-directed Research and Development program and by the U.S. Department of Energy, Office of Advanced Scientific Computing Research under the Collaboratory on Mathematics and Physics-Informed Learning Machines for Multiscale and Multiphysics Problems (PhILMs) project. SNL is a multimission laboratory managed and operated by National Technology and Engineering Solutions of Sandia, LLC., a wholly owned subsidiary of Honeywell International, Inc., for the U.S. Department of Energy's National Nuclear Security Administration under contract {DE-NA0003525}. This paper, SAND2022-0110, describes objective technical results and analysis. Any subjective views or opinions that might be expressed in this paper do not necessarily represent the views of the U.S. Department of Energy or the United States Government.

\appendix


\section{Detailed Numeric Results for Governing Law Learning Tasks}\label{sec:newapp_pde}

In this section we provide the detailed numerical results of governing law learning examples for different algorithms, as a continuation of the discussion in Section \ref{section:pde} of the main text and as the supplementary results of the training and test errors plotted in Figures \ref{fig:poisson_loss}, \ref{fig:loss_2ddarcy_16} and \ref{fig:loss_2ddarcy_s31} of the main text. The full results for 1D Poisson's equation learning, 2D Darcy's equation learning (from ``coarse'' training dataset) and 2D Darcy's equation learning (from ``fine'' training dataset) are provided in Tables \ref{tab:1DPoisson_new}, \ref{tab:2DDarcy} and \ref{tab:2DDarcy_reso_more}, respectively.

\begin{table}[]
\footnotesize
    \centering
{    \begin{tabular}{|c|>{\hspace{-4pt}}c<{\hspace{-4pt}}|>{\hspace{-4pt}}c<{\hspace{-4pt}}|>{\hspace{-4pt}}c<{\hspace{-4pt}}|>{\hspace{-4pt}}c<{\hspace{-4pt}}|>{\hspace{-4pt}}c<{\hspace{-4pt}}|>{\hspace{-4pt}}c<{\hspace{-4pt}}|}
    \hline\hline
    \multicolumn{7}{|c|}{Trained with normalized dataset}\\
    \cline{1-7}
   Model/dataset  &  $L=1$ &$L=2$&$L=4$&$L=8$&$L=16$&$L=32$\\
    \hline
GKN,train & 2.38e-2$\pm$1.45e-2 &  3.01e-2$\pm$0.27e-2 & 7.45e-1$\pm$1.92e-1  & 8.01e-1$\pm$1.98e-1  & 8.45e-1$\pm$1.54e-1 & 8.79e-1$\pm$1.20e-1\\
GKN,test& 2.22e-2$\pm$1.39e-2   & 3.05e-2$\pm$2.71e-3 & 7.47e-1$\pm$1.92e-1  & 8.03e-1$\pm$1.93e-1 & 8.38e-1$\pm$1.63e-1  &  8.78e-1$\pm$1.24e-1\\
\hline
    FNO,train &1.66e-2$\pm$2.03e-4 & 3.05e-3$\pm$5.82e-4 &  2.31e-3$\pm$2.58e-4&{\bf2.53e-3$\pm$1.97e-4} &  3.10e-3$\pm$6.97e-4 & 9.99e-1$\pm$8.11e-5 \\
    FNO,test & 2.07e-2$\pm$2.97e-4& 8.51e-3$\pm$4.11e-4& 1.14e-2$\pm$3.05e-4 & 9.71e-3$\pm$2.97e-4 & 1.60e-2$\pm$3.11e-4 & 1.00$\pm$8.05e-5 \\
\hline 
NKN,train& 1.05e-2$\pm$5.01e-4& 8.11e-3$\pm$3.78e-4 & 8.83e-3$\pm$5.11e-4 & 9.50e-3$\pm$4.95e-4  & 9.01e-3$\pm$1.97e-4  & 8.11e-3$\pm$3.95e-4 \\
NKN,test&1.22e-2$\pm$7.05e-4 &8.88e-3$\pm$5.97e-4 & 9.68e-3$\pm$4.71e-4& 1.02e-2$\pm$5.81e-4& 9.61e-3$\pm$3.12e-4& 8.60e-3$\pm$5.12e-4\\
\hline\hline
    \multicolumn{7}{|c|}{Trained with original (not normalized) dataset}\\
    \cline{1-7}
   Model/dataset  &  $L=1$ &$L=2$&$L=4$&$L=8$&$L=16$&$L=32$\\
    \hline
GKN,train&5.10e-3$\pm$3.07e-4 & 9.99e-1$\pm$8.11e-5 & 6.03e-3$\pm$2.43e-1 & 6.04e-1$\pm$2.43e-1 & 8.01e-1$\pm$1.98e-1 & 9.99e-1$\pm$8.11e-5\\
GKN,test&4.63e-3$\pm$5.17e-4&1.00$\pm$8.12e-5&6.04e-1$\pm$2.44e-1 & 6.05e-1$\pm$2.43e-1 & 8.03e-1$\pm$1.99e-1 & 1.00$\pm$1.42e-3\\
\hline 
FNO,train&2.14e-2$\pm$5.01e-4&1.01e-2$\pm$8.72e-4
& 3.51e-3$\pm$3.02e-5 & 3.32e-3$\pm$4.78e-5 & 4.11e-3$\pm$2.08e-4 & 9.99e-1$\pm$8.40e-5\\
FNO,test&2.70e-2$\pm$8.23e-4&6.48e-2$\pm$5.0e-1
&1.08e-1$\pm$3.50e-3 &8.47e-2$\pm$3.92e-3&2.48e-1$\pm$5.51e-2 & 1.00$\pm$8.11e-5\\
\hline
NKN,train& {\bf2.11e-3$\pm$1.62e-4} & {\bf 1.42e-3$\pm$5.98e-5} & {\bf1.94e-3$\pm$8.65e-5} & 2.71e-3$\pm$1.60e-4 & {\bf2.80e-3$\pm$2.14e-4} & {\bf3.60e-3$\pm$3.01e-4} \\
NKN,test&{\bf2.32e-3$\pm$1.71e-4}&{\bf1.50e-3$\pm$5.76e-5}& {\bf2.13e-3$\pm$9.88e-5} & {\bf2.83e-3$\pm$1.71e-4}&{\bf2.92e-3$\pm$2.21e-4}&{\bf3.90e-3$\pm$3.28e-4}\\
\hline 
    \end{tabular}}
\caption{{Leaning governing law example 1: 1D Poisson's equation. Relative mean squared errors (means $\pm$ standard errors) of the network predictions with respect to the reference solution (lower is better). Bold numbers highlight the case with the best error.}}
    \label{tab:1DPoisson_new}
\end{table}

\begin{table}[]
\scriptsize
    \centering
{    \begin{tabular}{|>{\hspace{-4pt}}c<{\hspace{-4pt}}|>{\hspace{-4pt}}c<{\hspace{-4pt}}|>{\hspace{-4pt}}c<{\hspace{-4pt}}|>{\hspace{-4pt}}c<{\hspace{-4pt}}|>{\hspace{-4pt}}c<{\hspace{-4pt}}|>{\hspace{-4pt}}c<{\hspace{-4pt}}|>{\hspace{-4pt}}c<{\hspace{-4pt}}|>{\hspace{-4pt}}c<{\hspace{-4pt}}|}
\hline\hline
    \multicolumn{8}{|c|}{Trained with normalized dataset}\\
    \cline{1-8}
   Model/dataset  &  $L=1$ &$L=2$&$L=4$&$L=8$&$L=16$&$L=32$&$L=64$\\
    \hline
GKN,train,s=16&  8.01e-2$\pm$1.88e-4&5.27e-2$\pm$1.44e-4&2.65e-2$\pm$3.83e-4&1.29e-2$\pm$3.53e-4&1.51e-2$\pm$2.62e-4&1.62e-2$\pm$5.49e-4 & 2.58e-2$\pm$5.63e-4\\
GKN,test,s'=16&  8.45e-2$\pm$1.56e-4&6.57e-2$\pm$9.70e-5&5.09e-2$\pm$6.02e-4&4.59e-2$\pm$1.20e-3&3.69e-2$\pm$9.28e-4&3.20e-2$\pm$9.62e-4& 3.97e-2$\pm$1.49e-3\\
GKN,test,s'=31& 8.66e-2$\pm$1.28e-4 & 6.97e-2$\pm$1.47e-4  &5.80e-2$\pm$5.08e-4 & 5.71e-2$\pm$9.66e-4 &5.21e-2$\pm$1.30e-3& 4.96e-2$\pm$1.16e-3 & 5.26e-2$\pm$2.33e-3\\
GKN,test,s'=61& 8.54e-2$\pm$1.44e-4& 6.79e-2$\pm$1.33e-4 & 5.58e-2$\pm$4.98e-4 & 5.35e-2$\pm$1.08e-3 &4.71e-2$\pm$9.66e-4 & 4.27e-2$\pm$7.63e-4 & 4.70e-2$\pm$2.23e-3\\
\hline
FNO,train,s=16& {\bf1.44e-2$\pm$5.5e-5} & {\bf8.88e-4$\pm$1.64e-5} & {\bf8.46e-4$\pm$2.17e-5} & {\bf1.09e-3$\pm$1.39e-4} & {\bf1.04e-3$\pm$2.53e-5} & 2.79e-1$\pm$8.11e-5 & 2.79e-1$\pm$8.43e-5 \\
FNO,test,s'=16& {\bf4.56e-2$\pm$2.37e-3} &  {\bf4.06e-2$\pm$9.97e-4} & 4.69e-2$\pm$8.88e-4 & 6.67e-2$\pm$1.07e-3 & 8.46e-2$\pm$1.03e-3 & 2.81e-1$\pm$8.12e-5 &2.81e-1$\pm$7.52e-5\\
FNO,test,s'=31& {\bf6.67e-2$\pm$1.50e-3} & {\bf4.59e-2$\pm$1.20e-3} & {\bf5.33e-2$\pm$1.60e-3} & 7.60e-2$\pm$9.01e-4 & 9.10e-2$\pm$2.40e-3&2.81e-1$\pm$8.15e-5&2.81e-1$\pm$8.38e-5\\
FNO,test,s'=61& {\bf7.02e-2$\pm$1.70e-3} &  {\bf4.82e-2$\pm$1.30e-3} & 5.56e-2$\pm$1.60e-3& 7.78e-2$\pm$8.10e-4 &9.29e-2$\pm$2.60e-3&2.81e-1$\pm$8.13e-5 & 
2.81e-1$\pm$8.39e-5\\
\hline
NKN,train,s=16&8.29e-2$\pm$1.96e-4& 5.90e-2$\pm$5.81e-4&3.44e-2$\pm$1.18e-3&1.38e-2$\pm$1.54e-3&7.98e-3$\pm$1.16e-3&{\bf6.89e-3$\pm$1.05e-3}& {\bf6.26e-3$\pm$1.09e-3}\\
NKN,test,s'=16&8.51e-2$\pm$4.07e-4&6.64e-2$\pm$1.26e-3&{\bf4.53e-2$\pm$1.80e-3}&{\bf2.18e-2$\pm$1.24e-3} & {\bf1.29e-2$\pm$7.61e-4}& {\bf1.14e-2$\pm$7.04e-4} & {\bf1.09e-2$\pm$4.21e-4}\\
NKN,test,s'=31&8.80e-2$\pm$2.99e-4&7.22e-2$\pm$1.16e-4&5.77e-2$\pm$8.30e-4&{\bf4.51e-2$\pm$1.88e-3} & {\bf3.99e-2$\pm$2.02e-3}& {\bf3.93e-2$\pm$2.35e-3}&{\bf3.91e-2$\pm$2.30e-3}\\
NKN,test,s'=61&8.64e-2$\pm$2.22e-4& 6.94e-2$\pm$3.70e-5&{\bf5.27e-2$\pm$5.30e-4} & {\bf3.77e-2$\pm$1.46e-3} & {\bf3.28e-2$\pm$7.36e-4} & {\bf3.23e-2$\pm$8.95e-4} & {\bf3.21e-2$\pm$1.09e-3}\\
\hline\hline
    \multicolumn{8}{|c|}{Trained with original (not normalized) dataset}\\
    \cline{1-8}
   Model/dataset  &  $L=1$ &$L=2$&$L=4$&$L=8$&$L=16$&$L=32$&$L=64$\\
    \hline
GKN,train,s=16&2.72e-1$\pm$1.04e-1&2.87e-1$\pm$9.59e-2&1.72e-1$\pm$4.56e-2&2.09e-1$\pm$3.50e-2 & 3.02e+2$\pm$1.77e+2&INF&INF\\
GKN,test,s'=16& 8.23e-1$\pm$1.71e-3& 5.04e-1$\pm$1.56e-1 & 2.32e-1$\pm$3.86e-2 & 5.00e-1$\pm$1.85e-1  & 1.07e+3$\pm$6.29e+2& INF&INF\\
GKN,test,s'=31& 7.95e-1$\pm$2.04e-1& 4.34e-1$\pm$1.08e-1 & 2.00e-1$\pm$4.26e-2 & 3.27e-1$\pm$6.88e-2  & 6.97e+2$\pm$3.84e+2& INF&INF\\
GKN,test,s'=61& 6.65e-1$\pm$1.54e-1&3.86e-1$\pm$1.02e-1  & 1.95e-1$\pm$4.15e-2 & 3.17e-1$\pm$4.18e-2  & 5.74e+2$\pm$3.28e+2& INF&INF\\
\hline
FNO,train,s=16& 2.89e-2$\pm$2.07e-4 & 7.02e-3$\pm$3.70e-4 & 5.11e-3$\pm$2.83e-4 & 4.90e-3$\pm$2.27e-4 & 4.50e-3$\pm$3.47e-4 & 5.17e-1$\pm$1.06e-1 & 6.35e-1$\pm$3.69e-5\\
FNO,test,s'=16& 6.04e-2$\pm$1.15e-3 & 6.95e-2$\pm$3.77e-3 &7.37e-2$\pm$5.27e-3 & 8.68e-2$\pm$2.33e-3  & 9.63e-2$\pm$2.16e-3 & 5.32e-1$\pm$ 8.71e-2  & 6.32e-1$\pm$1.24e-3\\
FNO,test,s'=31& 5.80e-1$\pm$1.56e-1 & 1.73e-1$\pm$1.97e-2 & 1.15e-1$\pm$1.59e-3 & 1.22e-1$\pm$2.43e-3 & 1.36e-1$\pm$4.70e-3 & 5.09e-1$\pm$8.48e-2 & 5.91e-1$\pm$1.27e-3 \\
FNO,test,s'=61& 6.96e-1$\pm$2.23e-1& 2.01e-1$\pm$2.07e-2&1.40e-1$\pm$9.34e-4&1.45e-1$\pm$2.60e-3 &1.56e-1$\pm$4.36e-3 & 4.98e-1$\pm$7.90e-2 & 5.72e-1$\pm$2.14e-3\\
\hline 
NKN,train,s=16& 9.10e-2$\pm$4.77e-4 & 8.68e-2$\pm$6.61e-4 & 8.45e-2$\pm$9.41e-4&8.21e-2$\pm$1.21e-3 & 8.04e-2$\pm$1.66e-3 & 7.67e-2$\pm$1.84e-3 & 7.41e-2$\pm$1.81e-3\\
NKN,test,s'=16&9.41e-2$\pm$4.96e-4&9.05e-2$\pm$7.12e-4&8.76e-2$\pm$1.01e-3&8.55e-2$\pm$1.11e-3 & 8.40e-2$\pm$1.72e-3& 8.02e-2$\pm$1.82e-3 & 7.76e-2$\pm$1.92e-3\\
NKN,test,s'=31&9.68e-2$\pm$4.74e-4&9.21e-2$\pm$8.57e-4&8.99e-2$\pm$9.89e-4&8.86e-2$\pm$1.02e-3 & 8.77e-2$\pm$1.58e-3& 8.46e-2$\pm$1.71e-3 & 8.25e-2$\pm$1.75e-3\\
NKN,test,s'=61& 9.47e-2$\pm$3.91e-4&8.96e-2$\pm$8.49e-4 &8.77e-2$\pm$1.01e-3 & 8.64e-2$\pm$1.08e-3 & 8.54e-2$\pm$1.64e-3 & 8.23e-2$\pm$1.76e-3&8.03e-2$\pm$1.79e-3\\
    \hline
    \end{tabular}}
    \caption{Leaning governing law example 2: 2D Darcy's equation with ``coarse'' training dataset ($\Delta x=1/(s-1)$, $s=16$). Relative training errors and errors on test datasets with different resolutions ($\Delta x=1/(s'-1)$, $s'\in\{16,31,61\}$) are provided. Relative mean squared errors (means $\pm$ standard errors) of the network predictions with respect to the reference solution (lower is better). Bold numbers highlight the case with the best error. ``INF'' denotes the cases where the final training loss is larger than $1e5$.}
    \label{tab:2DDarcy}
\end{table}

\begin{table}[]
\footnotesize
    \centering
{    \begin{tabular}{|>{\hspace{-4pt}}c<{\hspace{-4pt}}|>{\hspace{-4pt}}c<{\hspace{-4pt}}|>{\hspace{-4pt}}c<{\hspace{-4pt}}|>{\hspace{-4pt}}c<{\hspace{-4pt}}|>{\hspace{-4pt}}c<{\hspace{-4pt}}|>{\hspace{-4pt}}c<{\hspace{-4pt}}|>{\hspace{-4pt}}c<{\hspace{-4pt}}|}
\hline\hline
    \multicolumn{7}{|c|}{Trained with normalized dataset}\\
    \cline{1-7}
   Model/dataset  &  $L=1$ &$L=2$&$L=4$&$L=8$&$L=16$&$L=32$\\
\hline
GKN,train,s=31& 8.29e-2$\pm$1.94e-4&5.67e-2$\pm$1.63e-4&3.37e-2$\pm$2.71e-4&1.70e-2$\pm$3.22e-4&1.66e-2$\pm$8.06e-4&1.65e-2$\pm$8.21e-4\\
GKN,test,s'=16&8.99e-2$\pm$1.98e-3 &6.93e-2$\pm$2.11e-4&5.83e-2$\pm$5.45e-4&{\bf5.42e-2$\pm$1.25e-3} &5.01e-2$\pm$2.95e-3&5.41e-2$\pm$2.80e-3\\
GKN,test,s'=31& 8.29e-2$\pm$1.94e-4 & 6.82e-2$\pm$1.05e-4 & 5.41e-2$\pm$3.97e-4  & 4.57e-2$\pm$8.49e-4  & 3.66e-2$\pm$ 8.57e-4 &3.33e-2$\pm$2.14e-4 \\
GKN,test,s'=61& 8.92e-2$\pm$2.09e-3& 6.66e-2$\pm$8.50e-4 & 5.25e-2$\pm$4.13e-4 & 4.40e-2$\pm$7.57e-4  & 3.53e-2$\pm$1.19e-3& 3.59e-2$\pm$2.57e-3\\
\hline
FNO,train,s=31& {\bf1.42e-2$\pm$8.9e-5}  & {\bf1.31e-3$\pm$6.70e-5}  & {\bf8.62e-4$\pm$1.66e-5}  & {\bf9.31e-4$\pm$1.79e-5}  & {\bf1.02e-3$\pm$1.98e-5} &  2.79e-1$\pm$8.11e-5\\
FNO,test,s'=16& {\bf4.04e-2$\pm$3.12e-4}&{\bf4.61e-2$\pm$9.60e-4} & {\bf4.92e-2$\pm$6.18e-4}&6.58e-2$\pm$1.50e-3&8.21e-2$\pm$2.23e-3 & 2.81e-1$\pm$8.15e-5 \\
FNO,test,s'=31& {\bf3.58e-2$\pm$3.66e-4} & {\bf4.15e-2$\pm$1.12e-3} & 4.75e-2$\pm$7.56e-4 & 6.56e-2$\pm$1.30e-3 & 8.26e-2$\pm$2.35e-3 & 2.81e-1$\pm$8.12e-5  \\ 
FNO,test,s'=61&{\bf3.68e-2$\pm$3.40e-4}&{\bf4.18e-2$\pm$1.12e-3} & {\bf4.80e-2$\pm$7.55e-4}&6.58e-2$\pm$1.29e-3 &8.33e-2$\pm$2.34e-3 & 2.81e-1$\pm$8.12e-5 \\
\hline
NKN,train,s=31& 8.30e-2$\pm$4.57e-4 & 6.02e-2$\pm$1.44e-3&3.78e-2$\pm$1.57e-3&1.69e-2$\pm$1.91e-3 &8.38e-3$\pm$1.31e-3&{\bf9.56e-3$\pm$1.61e-3} \\
NKN,test,s'=16& 8.95e-2$\pm$2.86e-4 & 7.17e-2$\pm$8.02e-4 & 5.77e-2$\pm$9.27e-4 & 5.65e-2$\pm$4.73e-4 & {\bf4.14e-2$\pm$3.78e-3} &{\bf3.64e-2$\pm$2.40e-3}\\
NKN,test,s'=31&8.52e-2$\pm$2.05e-4&6.73e-2$\pm$2.32e-4&{\bf4.73e-2$\pm$5.97e-4}& {\bf2.33e-2$\pm$7.41e-4}&{\bf1.25e-2$\pm$6.30e-4}& {\bf1.31e-2$\pm$1.24e-3}\\
NKN,test,s'=61& 8.30e-2$\pm$8.00e-3& 6.55e-2$\pm$3.93e-4&4.57e-2$\pm$1.01e-3 & {\bf2.38e-2$\pm$7.59e-4} & {\bf1.38e-2$\pm$8.82e-4} & {\bf1.46e-2$\pm$1.35e-3}\\
\hline\hline
    \multicolumn{7}{|c|}{Trained with original (not normalized) dataset}\\
    \cline{1-7}
   Model/dataset  &  $L=1$ &$L=2$&$L=4$&$L=8$&$L=16$&$L=32$\\
\hline  
GKN,train,s=31&2.53e-1$\pm$9.50e-2 & 3.99e-1$\pm$7.44e-2&1.35e-1$\pm$4.81e-2&2.14e-1$\pm$1.93e-2&4.31e+2$\pm$2.51e+2&INF\\
GKN,test,s'=16& 1.78e+0$\pm$1.94e-1&1.01e+0$\pm$1.25e-1 &3.72e-1$\pm$3.54e-2 &1.20e+0$\pm$2.49e-1  & 3.78e+3$\pm$2.36e+3& INF\\
GKN,test,s'=31&1.17e+0$\pm$2.12e-1 & 7.02e-1$\pm$1.12e-1 & 1.97e-1$\pm$4.08e-2 &  3.39e-1$\pm$6.64e-2 & 1.40e+3$\pm$8.67e+2& INF\\
GKN,test,s'=61& 8.74e-1$\pm$1.42e-1&  5.92e-1$\pm$8.48e-2 & 1.81e-1$\pm$4.21e-2  & 2.99e-1$\pm$3.01e-2  &1.01e+3$\pm$5.87e+2 & INF \\
\hline
FNO,train,s=31&3.31e-2$\pm$9.27e-4 &1.07e-2$\pm$8.27e-4&5.07e-3$\pm$3.05e-4&4.96e-3$\pm$2.37e-4&5.43e-3$\pm$4.87e-4&1.32e-1$\pm$1.03e-1\\
FNO,test,s'=16& 9.82e-2$\pm$2.57e-3 & 1.12e-1$\pm$5.16e-3 & 1.05e-1$\pm$4.60e-3 & 1.14e-1$\pm$1.18e-3 & 1.36e-1$\pm$6.89e-3 & 3.49e-1$\pm$7.74e-2 \\
FNO,test,s'=31& 5.60e-2$\pm$1.74e-3 & 6.07e-2$\pm$4.42e-3&6.34e-2$\pm$3.28e-3&8.11e-2$\pm$1.89e-3 &9.19e-2$\pm$3.66e-3&2.65e-1$\pm$7.97e-2 \\ 
FNO,test,s'=61& 7.01e-2$\pm$3.71e-3 & 7.52e-2$\pm$4.84e-3 & 7.37e-2$\pm$3.81e-3 & 8.92e-2$\pm$2.15e-3 & 9.81e-2$\pm$4.17e-3 & 2.62e-1$\pm$8.54e-2 \\
\hline
NKN,train,s=31& 1.18e-1$\pm$1.99e-3& 9.38e-2$\pm$1.67e-3 & 9.16e-2$\pm$1.42e-3 & 9.21e-2$\pm$3.57e-3 & 8.82e-2$\pm$1.72e-3 &9.06e-2$\pm$4.21e-3\\
NKN,test,s'=16&1.25e-1$\pm$2.39e-3 & 1.02e-1$\pm$2.08e-3 & 9.91e-2$\pm$1.69e-3 & 9.92e-2$\pm$4.23e-3 & 9.48e-2$\pm$2.00e-3 & 1.04e-1$\pm$1.10e-3\\
NKN,test,s'=31&1.18e-1$\pm$2.09e-3 & 9.58e-2$\pm$1.60e-3 & 9.29e-2$\pm$1.37e-3 & 9.44e-2$\pm$3.67e-3 & 9.05e-2$\pm$1.80e-3 & 1.03e-1$\pm$1.14e-3\\
NKN,test,s'=61&1.15e-1$\pm$2.11e-3 &9.27e-2$\pm$1.39e-3 & 9.03e-2$\pm$1.24e-3 & 9.17e-2$\pm$3.49e-3 & 8.81e-2$\pm$1.85e-3 &1.02e-1$\pm$1.09e-3\\
\hline
    \end{tabular}}
    \caption{Leaning governing law example 2: 2D Darcy's equation with ``fine'' training dataset ($\Delta x=1/(s-1)$, $s=31$). Relative training errors and errors on test datasets with different resolutions ($\Delta x=1/(s'-1)$, $s'\in\{16,31,61\}$) are provided. Relative mean squared errors (means $\pm$ standard errors) of the network predictions with respect to the reference solution (lower is better). Bold numbers highlight the case with the best error. ``INF'' denotes the cases where the final training loss is larger than $1e5$.}
    \label{tab:2DDarcy_reso_more}
\end{table}

\bibliographystyle{elsarticle-num}
\bibliography{snl}

\end{document}